\documentclass{article}

\usepackage[preprint]{neurips_2023}

\usepackage[utf8]{inputenc} 
\usepackage[T1]{fontenc}    
\usepackage{hyperref}       
\usepackage{url}            
\usepackage{booktabs}       
\usepackage{amsfonts}       
\usepackage{nicefrac}       
\usepackage{microtype}      
\usepackage{xcolor}         
\usepackage{graphicx}       
\usepackage{algorithm, algorithmicx,algpseudocode}
\usepackage[title]{appendix}
\usepackage{wrapfig}
\usepackage{subcaption}
\usepackage{lipsum}
\usepackage{natbib}
\setcitestyle{numbers,square}

\title{Self-attention Dual Embedding for Graphs with Heterophily}

\author{
    Yurui Lai$^{*}$, Taiyan Zhang$^{*}$, Rui Fan\\
    School of Information Science and Technology, ShanghaiTech University, Shanghai, China\\
    \{laiyr, zhangty2022, fanrui\}@shanghaitech.edu.cn
}

\begin{document}
\newcommand{\indep}{\perp\!\!\!\perp}
\newcommand{\notindep}{\not\!\perp\!\!\!\perp}

\maketitle
\newcommand\blfootnote[1]{%
\begingroup
\renewcommand\thefootnote{}\footnote{#1}%
\addtocounter{footnote}{-1}%
\endgroup
}
\blfootnote{$^{*}$Equal contribution\\}

\begin{abstract}
    Graph Neural Networks (GNNs) have been very successful in the node classification task. GNNs typically assume graphs to be homophilic, i.e. neighboring nodes are likely to belong to the same class. However, a number of real-world graphs are heterophilic, which leads to much lower classification accuracy using standard GNNs. In this work, we design a novel GNN that is effective for both heterophilic and homophilic graphs. Our work is based on three main observations. First, we show that node features and graph topology provide different amounts of informativeness in different graphs, and therefore they should be encoded independently and prioritized in an adaptive manner. Second, we show that allowing negative attention weights when propagating graph topology information improves accuracy. Finally, we show that asymmetric attention weights between nodes are helpful. We design a GNN that makes use of these observations through a novel self-attention mechanism. We evaluated our algorithm on various real-world graphs and showed that we achieve state-of-the-art results compared to existing GNNs. We also analyze the effectiveness of the main components of our design on different graphs.
\end{abstract}

\section{Introduction}
\label{sec:intro}

The graph node classification task tries to predict the classes of nodes on the basis of their features and the graph topology. An edge between two nodes indicates the correlation between the classes of the nodes. Many real-world graphs are homophilic\cite{sen2008collective}, meaning there is positive correlation between an edge's endpoints. However, other graphs are naturally heterophilic, so that edges, while still informative, often indicate negative correlation between the endpoints\cite{zhu2020beyond}. For example, in the University of Wisconsin web graph, there are frequently edges between different classes of web pages, for example from a faculty's homepage to web pages of their students and courses\cite{peigeom}. 

Graph Neural Network(GNN) achieves great success on node classification\cite{kipf2016semi} \cite{velivckovicgraph}\cite{xu2018representation}\cite{wu2019simplifying}\cite{yang2021rethinking}. However, standard GNNs implicitly assume graphs are homophilic, because the aggregation step in most GNNs pushes the embeddings of neighboring nodes close to each other, leading to neighbors often receiving the same class. When such GNNs are applied to heterophilic graphs, their accuracy can degrade substantially\cite{peigeom}\cite{zhu2020beyond}\cite{yang2022graph}. 

A number of recent works\cite{peigeom}\cite{zhu2020beyond}\cite{chien2020adaptive}\cite{bo2021beyond}\cite{yang2021diverse}
\cite{yan2022two}\cite{luanrevisiting}\cite{li2022finding}\cite{fang2022polarized} focus on improving GNN accuracy on heterophilic graphs . We divide these works into two types, depending on whether they make use of graph topology information in an implicit or explicit manner. To overcome the problems of traditional GNNs, many methods in the first type allow negative attention weights between nodes during aggregation\cite{chien2020adaptive}\cite{bo2021beyond}\cite{yang2021diverse}. Viewing the nodes' classes as a graph signal, these algorithms apply both high and low pass filters on the signal, whereas standard GNNs use only positive attention weights and thus only apply low pass filters. Besides, other techniques such as multi-hop embedding combination \cite{jin2021universal}, node sequence sampling\cite{fang2022polarized}, etc. have been used in this type. The first type of algorithm implicitly makes use of the graph topology via message passing along edges. 
The second type of heterophilic GNNs also use negative attention weights, and additionally make explicit use of topology information by treating each node's adjacency vector, giving its connections to all other nodes, as an additional feature\cite{yan2022two}\cite{li2022finding}\cite{luanrevisiting}\cite{yang2022graph}. It embeds nodes using a combination of their standard and adjacency features. This approach is sometimes highly effective and can substantially improve classification accuracy. 

In this work, we propose several novel techniques that further increase the accuracy of heterophilic graphs. We first observe that node and topology features provide different degrees of informativeness in different graphs. Thus, when one type of feature is substantially more informative than the other, we should avoid premature mixing of the features to avoid "contamination". Second, we show that using negative attention weights when embedding topology features improves accuracy, in the same way, that such weights are useful when embedding node features. Finally, while existing heterophilic GNNs use only symmetric edge weights, we show that it is beneficial to allow asymmetric attention weights between nodes, i.e. to let one node i influence node j more than j influences i thought edges $(i,j)$ and $(j,i)$ with different weights.

Based on these observations, we propose a novel GNN model which we call SADE-GCN (self-attention dual embedding graph convolutional network). SADE-GCN explicitly separates the embeddings of the node and topology features and only combines them in the final step of the model. It allows negative attention weights when computing both node and topology embeddings. Furthermore, it uses a novel self-attention mechanism to enable asymmetric attention weights. We give an efficient implementation of our algorithm on a range of benchmark graphs. Our algorithm achieves high accuracy on homophilic graphs and is more accurate than all other GNNs on nearly all heterophilic graphs. We conduct a number of ablations to validate different aspects of our design. Results show that the unique aspects of our model often substantially improve accuracy.

In summary, our contributions are as follows.
\begin{itemize}
\item We observe that the importance of topology and feature varies among graphs, and show that using negative attention weights when embedding topology features improves accuracy for heterophilic graphs. Additionally, we highlight the benefits of using asymmetric attention weights between neighbors.

\item To leverage our observations, we propose a new model called SADE-GCN. This model is based on Dual Embedding architecture that keeps the independence of feature and topology during message passing. Furthermore, SADE-GCN incorporates a novel self-attention mechanism that extracts the heterophily and asymmetry of the graph. 

\item Finally, we assess the effectiveness of SADE-GCN on a total of 9 graphs, including 6 heterophilic and 3 homophilic ones. We demonstrate that our proposed model outperforms other existing state-of-the-art methods in terms of prediction accuracy. Additionally, we investigate the contributions of individual modules by conducting ablation experiments. 
\end{itemize}

\section{Preliminaries}
In this section, we introduce notation and background knowledge. A graph $\mathcal{G} = (\mathcal{V}, \mathcal{E})$, $\mathcal{V}$ is the node set and $\mathcal{E}$ is edge set. There are $N$ nodes and $E$ edges in a graph. We denote graph adjacency matrix as $\mathbf{A}\in \mathbb{R}^{N\times N}$. The degree matrix of the graph is $\mathbf{D}\in \mathbb{R}^{N\times N}$. The node feature is $\mathbf{X}\in \mathbb{R}^{N\times F}$ and the node label is $\mathbf{Y}\in \mathbb{R}^{N\times C}$, where $C$ is the number of classes. 

\subsection{Graph Neural Network}
Traditional GNNs rely on the message passing mechanism, which entails two phases: aggregation and transformation. During the aggregation phase, the embeddings are propagated along an adjacency matrix. Then, embeddings after propagation are transformed by non-linear or linear layers.
The mechanism can be recursively defined as follows. 
\begin{equation}
\label{e1}
    \mathbf{H}^{(0)}=\mathbf{X}, \qquad \mathbf{H}^{(l)}=\sigma(\mathbf{\tilde{A}}\mathbf{H}^{(l-1)}\mathbf{W}^{(l)})
\end{equation}
where the hidden feature embedding of $l$ layer of GNN is $\mathbf{{H}}^{l}\in \mathbb{R}^{N\times F^{(l)}}$ and $\sigma$ is non-linear function.  $\mathbf{\tilde{A}}$ represents $ {\mathbf{D}}^{-1}\mathbf{A}$, ${\mathbf{D}}^{-1/2}\mathbf{A}\mathbf{D}^{-1/2}$, etc. Since the normalized adjacency is a low-pass filter\cite{bo2021beyond}, the embeddings of neighboring nodes tend to become more similar. When the heterophily is high, nodes of different classes often have similar embeddings, which subsequently leads to a significant reduction in GNN performance.

\subsection{Self-attention in Transformer}
The transformer was first proposed in natural language processing \cite{vaswani2017attention} and has since gained popularity with great success in other fields, such as computer vision \cite{dosovitskiy2020image}\cite{he2022masked} and bioinformatics \cite{senior2020improved}\cite{jumper2021highly}. Recently, transformers have also achieved success on graph applications \cite{yun2019graph}\cite{zhang2020graph}\cite{ying2021transformers}\cite{rampavsek2022recipe}.  The fundamental element of graph transformers is self-attention. Let hidden embedding of $l$ layer of transformer be $\mathbf{{H}}^{(l)}\in \mathbb{R}^{N\times H^{(l)}}$. 
For layer $l$, let query weight, key weight and value weight be ${\mathbf{{W}}_{Q}}^{(l)}$, ${\mathbf{{W}}_{K}}^{(l)}$,${\mathbf{{W}}_{V}}^{(l)}$ $\in \mathbb{R}^{H^{(l-1)}\times H^{(l)}}$ respectively, then self-attention can be expressed as 
\begin{equation}
\label{e3}
   \mathbf{Q}^{(l)} = (\mathbf{H}^{(l-1)}{\mathbf{{W}}_{Q}}^{(l)}), \qquad\mathbf{K}^{(l)} = (\mathbf{H}^{(l-1)}{\mathbf{{W}}_{K}}^{(l)}), \qquad  \mathbf{V}^{(l)} = (\mathbf{H}^{(l-1)}{\mathbf{{W}}_{V}}^{(l)}), 
\end{equation}
\begin{equation}
\label{e5}
    \mathbf{H}^{(l)}=Softmax( \frac{(\mathbf{Q}^{(l)}(\mathbf{K}^{(l)})^{\mathrm{T}}}{ \sqrt{H^{(l)}}} )(\mathbf{V}^{(l)}) + Res(\mathbf{H}^{(0)})
\end{equation}

Here, $Res()$ denotes the residual connection function\cite{He_2016_CVPR}\cite{vaswani2017attention}, and ${\mathbf{H}}^{(0)}$ denotes the embedding of the graph. The self-attention mechanism calculates the pairwise inner product between the input elements from the query ${\mathbf{Q}}^{l-1}$ and the key ${\mathbf{K}}^{l-1}$ matrices. The output is obtained by taking the weighted average of ${\mathbf{V}}^{l-1}$.

\begin{figure}[H]
    \begin{minipage}{0.5\linewidth}
        \centering
        \includegraphics[width=0.95\linewidth]{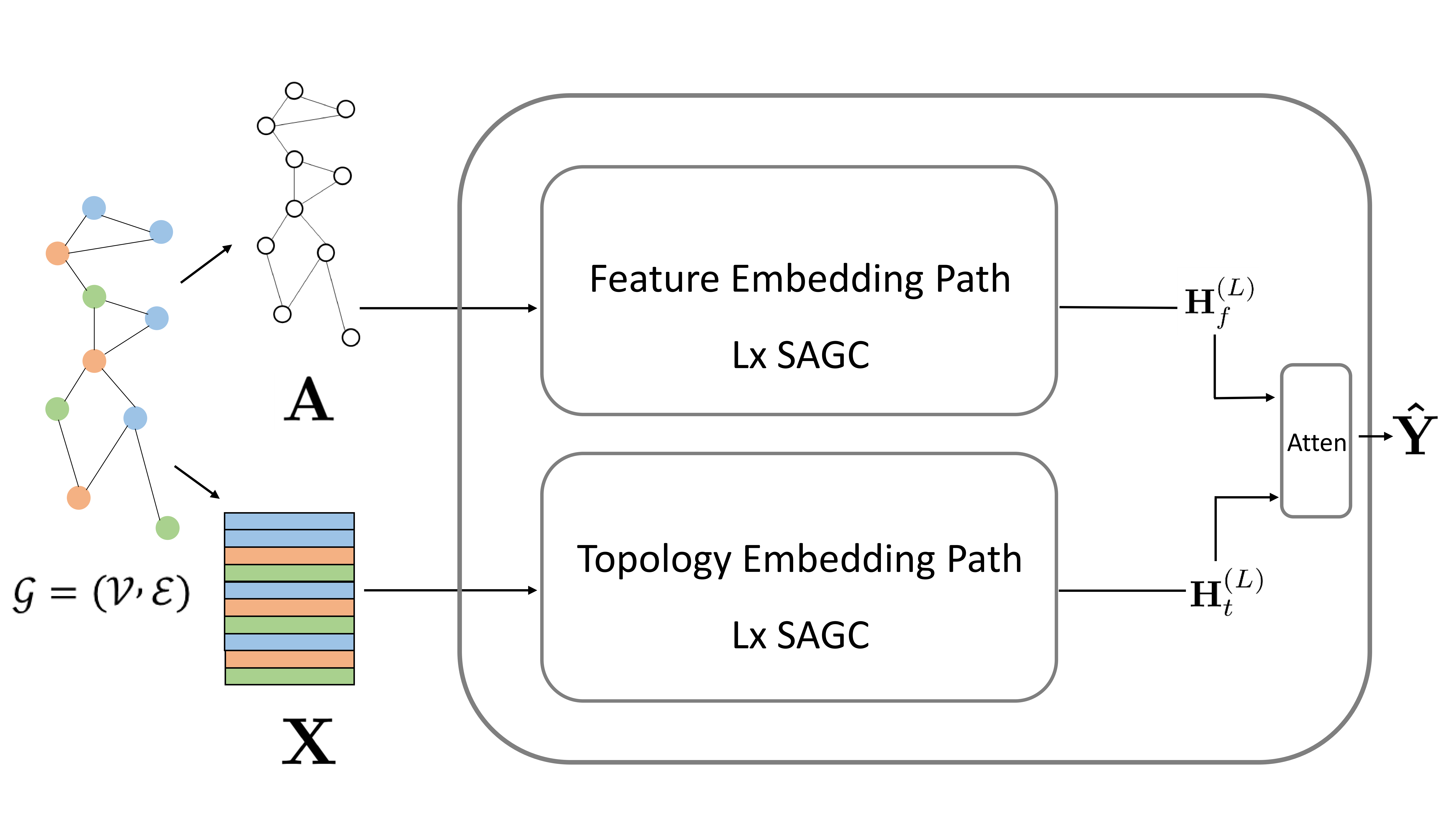}
    \subcaption{The overview of SADE-GCN}
    \end{minipage}
    \label{fig-SADE}
    \begin{minipage}{0.5\linewidth}
        \includegraphics[width=0.95\linewidth]{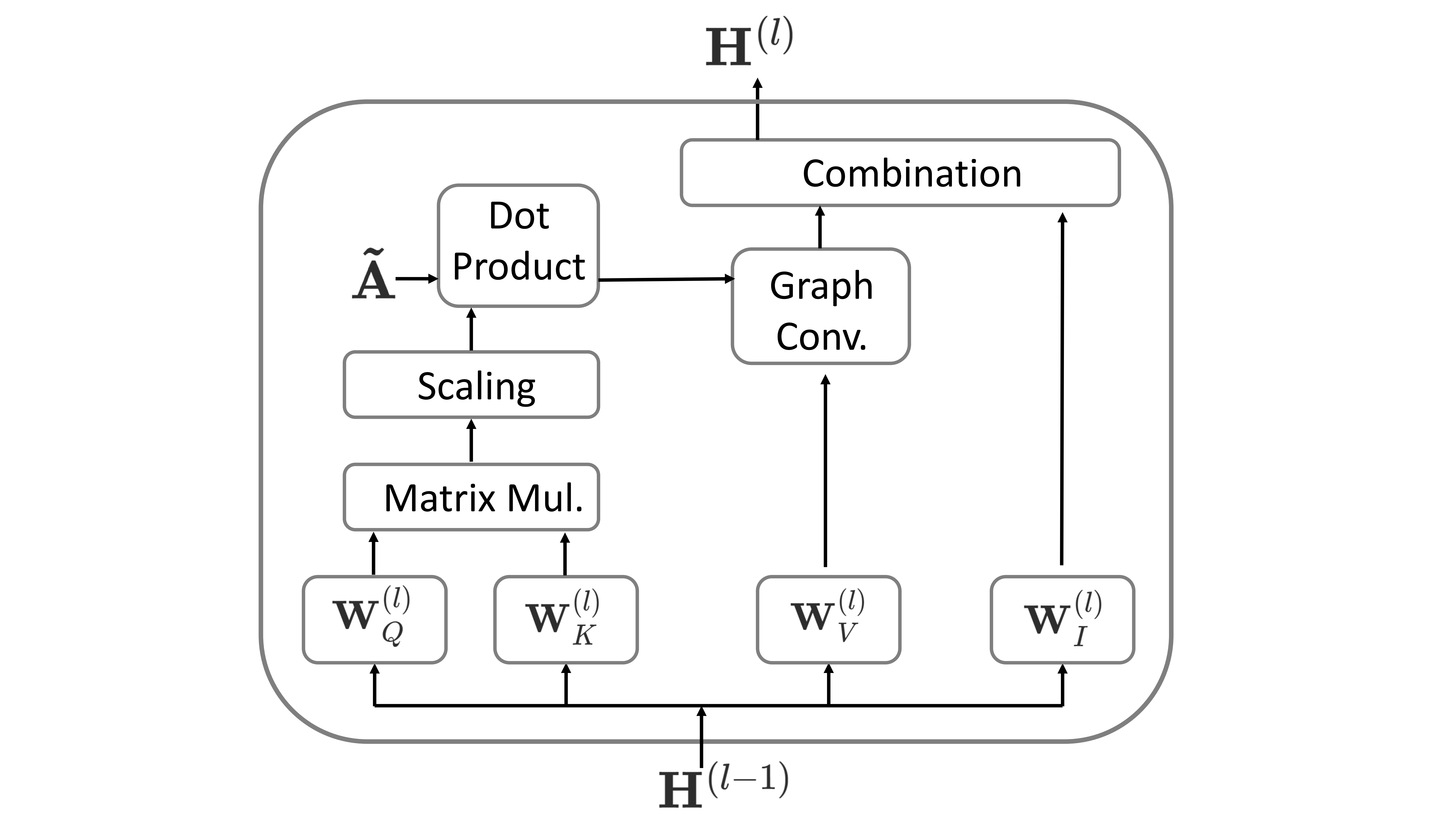}
        \subcaption{The details of SAGC in SADE-GCN}
    \end{minipage}
    \caption{Our method}
    \label{fig-SAGC}
\end{figure}

\section{Self-attention Dual Embedding}
This section presents an overview of our proposed method, SADE-GCN (self-attention dual embedding graph convolutional network), which is based on our three observations of heterophilic graphs and detailed analysis.

\subsection{Observations of Heterophilic Graphs}
\subsubsection{Different Importance of Feature and Topology}

As recent research \cite{yang2022graph} demonstrates, heterophilic graphs require a compromise between graph topology and node features. It is thus crucial to explicitly encode both feature and topology information in order to achieve effective performance. To that end, some existing GNN approaches for heterophilic graphs, such as \cite{yan2022two}, \cite{li2022finding}, and \cite{luanrevisiting}, emphasize the importance of explicitly embedding topology information and then combining it with feature embedding during message passing.

To evaluate the relative importance of feature and topology information in graphs, we conducted experiments for graph node classification using both MLP and LINK\cite{zheleva2009join}. MLP solely relies on feature information, while LINK is a model with a linear layer that takes the adjacency matrix as input, with each row of the matrix representing the topology of the corresponding node. MLP outperforms LINK on some heterophilic graphs, such as Wisconsin, achieving an accuracy of 85.29\% compared to 51.96\% with LINK, indicating that feature information is more important in those cases. However, on other graphs, including Chameleon, MLP underperforms relative to LINK, achieving an accuracy of only 46.21\% compared to 70.57\% using LINK, demonstrating the greater importance of topology information in those cases. The full results of these experiments are presented in Tables \ref{main-t1} and \ref{main-t2}. These results suggest that the relative importance of feature and topology information varies among graphs. We hypothesize that the different levels of noise in feature and topology may correspond to their different importance. Thus, we propose keeping the independence of feature and topology embeddings during message passing as a way to enhance the model's robustness to noise and boost overall performance. This approach is consistent with some prior work that acknowledges the presence of inherent noise in both feature and topology information \cite{zhao2021data}\cite{wu2022recent}.

\subsubsection{Graph Topology for Heterophily}

 From the perspective of graph signal processing, the raw adjacency matrix is a low-frequency filter that only captures the homophily of a graph \cite{bo2021beyond}\cite{luanrevisiting}. To account for heterophily, GNNs leverage high-frequency filters by introducing negative edge weights to the adjacency matrix. Specifically, when adjacent nodes with different class labels are connected by negative weights, the distance between the feature embeddings of these nodes increases.

 Previous studies have emphasized the role of feature embeddings in capturing heterophily \cite{bo2021beyond}\cite{luanrevisiting}. However, for effective extraction of heterophily, it is crucial to make use of both topology and feature information. In order to achieve this, we propose employing attention matrices for feature and topology embeddings respectively.

\subsubsection{Asymmetry of Heterophilic Graph}

In reality, neighbors in a heterophilic graph often possess varying influences over each other. For example, in fraudulent networks\cite{dou2020enhancing}, fraudsters often establish connections with regular users. Regular users tend to assume that fraudsters are also regular users, while fraudsters are unlikely to make the same mistake. Therefore, it's necessary to capture the asymmetry in influence. 

Traditional GNNs like GCN usually convert the input to a symmetric undirected graph. In \cite{yan2022two} \cite{luanrevisiting}\cite{he2022block}, an attention matrix representing the interaction between neighbors is used to adjust the adjacency matrix for message passing. The calculation of the attention matrix typically involves the computation of similarity measures such as cosine distance and inner product, which by design, are symmetric. An attention matrix with greater expressiveness and asymmetry is needed to more accurately adjust the weights of edges in the network.

\subsection{Dual Embedding Paths}

SADE-GCN consists of two parallel paths to embed features and topology independently. One is $\mathbf{P}_{f}$ which explicitly embeds features. It takes node features as input. Similarly, $\mathbf{P}_{t}$ captures topology information explicitly. There exist several techniques for embedding graph topology \cite{luanrevisiting}\cite{xu2022hp}\cite{dwivedi2020generalization}. In this study, we adopt the adjacency matrix as the input for $\mathbf{P}_{t}$. Each row of the adjacency matrix represents the topology of a node. Both paths through the network consist of SAGC (self-attention graph convolution) modules, which are introduced in the next section.

\subsection{Self-attention Graph Convolution}

\subsubsection{Self-attention for Graphs with Heterophily}
Firstly, we use self-attention to obtain the attention matrix. Self-attention has strong expressive power \cite{likhosherstov2021expressive}\cite{perez2019turing}\cite{perez2021attention}\cite{kim2022pure}, making it well-suited for capturing complex patterns within a graph network. To the best of our knowledge, this is the first application of self-attention in heterophilic graphs.

For $\mathbf{P}_{f}$ , the query ${\mathbf{W}_{Q_f}}^{(l)}$ and key weight ${\mathbf{W}_{K_f}}^{(l)}$ $\in \mathbb{R}^{H^{(l-1)}\times H_{K_f}^{(l)}}$. For $\mathbf{P}_{t}$ , the query ${\mathbf{W}_{Q_t}}^{(l)}$ and key weight ${\mathbf{W}_{K_t}}^{(l)}$ $\in \mathbb{R}^{H^{(l-1)}\times H_{Q_f}^{(l)}}$ . $\mathbf{P}_{f}$ and $\mathbf{P}_{t}$ have the same embedding mechanism. We use a unified framework to describe the process.

We can get the query and key for feature and topology in equation \ref{e6}. The attention matrix with positive value $ {\mathbf{R}_{f}}^{(l)} $ and ${\mathbf{R}_{t}}^{(l)}$ are respectively calculated by matrix multiplication of query and key in equation \ref{e8}.
\begin{equation}
\label{e6}
    \mathbf{Q}^{(l)} = Softmax(\mathbf{H}^{(l-1)}{\mathbf{W}_{Q}}^{(l)}) ,\qquad  \mathbf{K}^{(l)} = Softmax(\mathbf{H}^{(l-1)}{\mathbf{W}_{K}}^{(l)})
\end{equation}
\begin{equation}
\label{e8}
    \mathbf{R}^{(l)}= \mathbf{Q}^{(l)}(\mathbf{K}^{(l)})^{\mathrm{T}}
\end{equation} 

Here we introduce modifications to the standard self-attention mechanism. Specifically, we perform a Softmax operation on the query and key vectors prior to multiplication. This modification is necessary because the graphs are often large and sparse, we only need to focus on neighboring nodes. If we perform a Softmax operation on each row of the attention matrix, as in the standard self-attention mechanism, the attention weights for neighboring nodes will be excessively small due to interference from a large number of non-neighbor nodes.

The values of $ \mathbf{R}^{(l)}$ generated through our SADE-GCN model exist within the range of $[0,1]$. However, in order to accurately capture the heterophily of the graph network, it is necessary to include negative attention weights. To achieve this, we employ a scaling function, as defined in equation \ref{e9}, to map the values of $ \mathbf{R}^{(l)}$ from the range of $[0,1]$ to $[-1,1]$. Subsequently, we utilize the attention matrices to reweight the edge weights through element-wise multiplication, also known as the Hadamard Product equation \ref{e10}. Through this operation, our model selectively accounts only for neighboring nodes, thereby setting all non-neighbor attention weights to zero. In this way, we obtain $\mathbf{\tilde{A}}_{f}^{(l)}$ and $\mathbf{\tilde{A}}_{t}^{(l)}$ values for the feature and topology paths, respectively.
\begin{equation}
\label{e9}
    \mathbf{\hat{R}}^{(l)}=  2\mathbf{R}^{(l)}-1
\end{equation}
\begin{equation}
\label{e10}
    \mathbf{\tilde{A}}^{(l)} = \mathbf{\hat{R}}^{(l)} \odot \mathbf{\tilde{A}}
\end{equation}
We provide an efficient implementation to obtain the reweighted adjacency matrix $\mathbf{\tilde{A}}$. Details of this implementation can be found in the Appendix. In summary, our approach directly calculates the attention weights of neighbors instead of removing the non-neighbors after applying self-attention.

\subsubsection{Heterophilic Graph Convolution}
We give heterophilic graph convolution after getting the reweighted adjacency matrix. For feature, the value weight is ${\mathbf{W}_{V_f}}^{(l)}$ and residual weight is ${\mathbf{W}_{I_f}}^{(l)}$ $\in \mathbb{R}^{H^{(l-1)} \times H^{(l) }} $. For topology , the value weight is ${\mathbf{W}_{V_t}}^{(l)}$ and residual weight is ${\mathbf{W}_{I_t}}^{(l)}$ $\in \mathbb{R}^{H^{(l-1)} \times H^{(l) }} $. 
\begin{equation}
\label{e11}
     \mathbf{H}^{(l)}={\alpha_V}\mathbf{c}_{V}ReLU(\mathbf{\tilde{A}}^{(l)}\mathbf{H}^{(l-1)}{\mathbf{W}_{V}}^{(l)}) +   {\alpha_I}\mathbf{c}_{I}\mathbf{H}^{(l-1)}{\mathbf{W}_{I}}^{(l)}
\end{equation}
\begin{equation}
\label{e12}
    \mathbf{c}_{V}, \mathbf{c}_{I} = Softmax(\sigma(\mathbf{H}^{(l)}\mathbf{G}_{V}), \sigma(\mathbf{H}^{(l)}\mathbf{G}_{I}))
\end{equation}
where $\mathbf{G}_{V}$ and $\mathbf{G}_{I}$ are learnable weights,  ${\alpha_V}$ and ${\alpha_I}$ are hyperparameters used for manually weighing the message passing and residual result. $\sigma$ here is a sigmoid layer.

\subsubsection{Feature and Topology Embedding Combination}

The final output $\mathbf{\hat{Y}} \in \mathbb{R}^{N \times C}$ comes from an adaptive combination of two paths output $\mathbf{H}_{f}^{(L)}$ and $\mathbf{H}_{t}^{(L)}$, where $L$ represents length of the embedding path. 
\begin{equation}
\label{e14}
     \mathbf{\hat{Y}} =  {\alpha_f}\mathbf{c}_{f}{\mathbf{H}_{f}}^{(L)} +  {\alpha_t}\mathbf{c}_{t}\mathbf{H}_{t}^{(L)}
\end{equation}
where $\mathbf{c}_{f}$ and $\mathbf{c}_{t}$ are attention vectors, which are calculated in the same way as the above method. $ {\alpha_f}$ and $ {\alpha_t}$ are hyperparameters used for manually weighing features and topology embedding.

\section{Experiment} 
In this section, we first present our experimental settings. Next, we evaluate our method for node classification on several real-world graph datasets and compare it with state-of-the-art (SOTA) methods. Finally, we analyze the components of our method, as well as confirm the observations and analysis presented in the previous section.

\subsection{Settings}

\subsubsection{Datasets}

 We adopt a total of 9 real-world graph datasets in our experiments. The Cora, Citeseer, and PubMed datasets\cite{sen2008collective} are citation networks with high homophily. In \cite{peigeom}, a series of heterophilic graphs are proposed, including the Texas, Wisconsin, Cornell, Wikipedia networks Squirrel and Chameleon, and Actor co-occurrence network. These nine datasets are commonly used in research on heterophilic graphs. These graphs vary in feature dimension, homophily level, and number of nodes and edges, enabling us to fully test the ability of our model. We provide the characteristics of these graphs in Table \ref{g-info}, with additional descriptions available in the Appendix.

\begin{table*}[!htbp]
\caption{The overview of 9 graph datasets' characteristics. The edge homophily is measured by the ratio of the edges that connect nodes with the same label.}
\centering
\resizebox{\linewidth}{!}
{
\Huge
\begin{tabular}{c c c c c c c c c c}
\hline
    & \textbf{Texas} & \textbf{Wisconsin}     & \textbf{Cornell} & \textbf{Film} & \textbf{Squirrel} & \textbf{Chameleon} & \textbf{Cora} & \textbf{Citeseer} &\textbf{Pubmed}  \\
\textbf{Nodes}  &   183    &   251        &  183  &  7,600  & 5,201 & 2,277  & 2,708 & 3,327 &  19,717  \\
 \textbf{Edges}  &    295   &    466       &  280  &  26,752  & 198,493 & 31,421  &  5,278 & 4,676  &  44,327    \\
 \textbf{Features}  &   1,703    &    1,703       &  1,703   &  931  &  2,089 &  2,325 &  1,433 & 3,703  & 500  \\
 \textbf{Classes} &    5   &     5      & 5  &  5  & 5 & 5  & 6 & 7  &  3   \\
 \textbf{ Edge Hom.}  &   0.11    &      0.21     &  0.30  &  0.22  & 0.22  &  0.23 &  0.81 &  0.74 &  0.80  \\
 \hline  
 \end{tabular}
 }
 \label{g-info}
 \end{table*}

\begin{table*}[!htbp]
\caption{The overall accuracy results of the compared methods on the 9 small datasets. We highlight the best score on each dataset in bold and the second is underlined. Note that the error bar $(\pm)$ indicates the standard deviation score of the results over.}
    \centering
    \resizebox{\linewidth}{!}{
    \Huge
    \begin{tabular}{c c c c c c c c c cc}
    \hline
    & \textbf{Cora} & \textbf{Citeseer} & \textbf{Pubmed} & \textbf{Squirrel} & \textbf{Chameleon} & \textbf{Texas} & \textbf{Wisconsin}  & \textbf{Cornell} & \textbf{Film} & \textbf{Avg-Rank}\\
    \hline
    MLP & $75.69\pm 2.00$ & $74.02\pm 1.90$& $87.16\pm0.37 $ &$28.77\pm1.56 $ &$46.21\pm2.99 $ & $80.81\pm4.75 $& $85.29\pm3.31 $&$81.89\pm6.40 $ &$36.53\pm0.70 $ & $13.67$ \\
    LINK & $77.46\pm1.94 $ & $66.16\pm2.92 $& $78.74\pm0.47 $ &$62.31\pm1.78 $ &$70.57\pm2.00 $ & $58.10\pm6.75 $& $51.96\pm5.63 $&$50.54\pm5.92 $ &$24.11\pm0.70 $ & $15.22$ \\
    \hline
    GCN  & $86.89\pm1.27 $ &$76.50\pm1.36 $ &$88.42\pm0.50 $ &$53.43\pm2.01 $ &$64.82\pm2.24 $ & $55.14 \pm 5.16$    &  $51.76 \pm 3.06$  & $60.54 \pm 5.30$ & $27.32\pm 1.10$ & $14.56$   \\
    GAT     & $87.30\pm1.10 $&$76.55\pm1.23 $ & $86.33\pm0.48 $ & $40.72\pm1.55 $&$60.26\pm2.50 $ &  $52.16 \pm 6.63$   &  $49.41 \pm 4.09$       &$61.89 \pm 5.05$ & $27.44 \pm 0.89$& $15.89$  \\
    GraphSAGE     &$86.90\pm1.04 $ & $76.04\pm1.30 $& $88.45\pm0.50 $& $41.61\pm0.74 $& $58.73\pm1.68 $& $82.43 \pm 6.14$  &  $81.18 \pm 5.56$       & $75.95\pm 5.01$& $34.23\pm 0.99$ & $13.33$   \\
    MixHop & $87.61\pm0.85 $& $76.26\pm1.33 $& $85.31\pm0.61 $& $43.80\pm1.48 $& $60.50\pm2.53 $& $77.84 \pm 7.73$   &   $ 75.88\pm 4.90$      & $73.51 \pm 6.34$ & $ 32.22\pm 2.34$   & $14.11$ \\
    GCNII &  $ \mathbf{88.37\pm1.25} $& $77.33\pm1.48 $& $\mathbf{90.15\pm0.43} $&$38.47\pm1.58 $ & $63.86\pm3.04 $& $77.57 \pm 3.83$  &   $80.39 \pm 3.40$          & $77.86 \pm 3.79$ &  $37.44 \pm 1.30$ & $8.11$ \\
    \hline
    H$_2$GCN &$87.87\pm1.20 $ &$77.11\pm1.57 $ &$89.49\pm0.38 $ &$36.48\pm1.86 $ &$60.11\pm2.15 $ & $84.86 \pm 7.23$  & $87.65 \pm 4.98$ & $82.70 \pm 5.28$ & $35.70 \pm 1.00$  & $9.67$ \\
    FAGCN &$88.05\pm1.57 $ &$77.07\pm2.05 $ &$88.09\pm1.38 $ &$30.83\pm0.69 $ &$46.07\pm2.11 $ & $76.49 \pm 2.87$  & $79.61 \pm 1.58$ & $76.76 \pm 5.87$ & $34.82 \pm 1.35$  & $13.22$ \\
    GPR-GNN &$87.95\pm1.18 $&$77.13\pm1.67 $&$87.54\pm0.38$ & $46.31\pm2.46 $&$62.59\pm2.04 $& $81.35\pm 5.32$   &     $82.55 \pm 6.23$       & $78.11 \pm 6.55$ & $35.16 \pm 0.90$ & $10.67$ \\
    WRGAT &$88.20\pm2.26 $&$76.81\pm1.89 $&$88.52\pm0.92 $&$48.85\pm0.78 $&$65.24\pm0.87 $ & $83.62 \pm 5.50 $    &    ${ 86.98 \pm 3.78}$        & $ 81.62 \pm 3.90$ & $ 36.53 \pm 0.77 $ & $8.44$ \\
    ACM-GCN &$87.91\pm0.95 $&$77.32\pm1.70 $&$90.00\pm0.52 $&$54.40\pm1.88 $&$66.93\pm1.85 $& $ \underline{87.84  \pm 4.40}$   &     $  \underline{88.43 \pm 3.22}$       & $ 85.14 \pm 6.07 $ & $ 36.28 \pm 1.09 $ & $5.78$ \\
    \hline
    Geom-GCN &$85.35\pm1.57 $ &$\mathbf{78.02\pm1.15} $ &$89.95\pm0.47 $ &$38.15\pm0.92 $ &$60.00\pm2.81 $ & $66.76 \pm 2.72$  & $64.51 \pm 3.66$ & $60.54 \pm 3.67$ & $31.59 \pm 1.15$  & $13.33$\\
    LINKX &$87.86\pm0.77 $&$73.19\pm0.99 $&$87.86\pm0.77 $&$61.81\pm1.80 $&$68.42\pm1.38 $& $ 74.60  \pm 8.37$   &     $  75.49 \pm 5.72$       & $ 77.84 \pm 5.81 $ & $ 36.10 \pm 1.55 $ & $11.89$\\
    GGCN &$87.95\pm1.05 $&$77.14\pm1.45 $&$89.15\pm0.37 $&$55.17\pm1.58 $&$71.14\pm1.84 $& $ 84.86  \pm 4.55$   &     $  86.86 \pm 3.29$       & $ 85.68 \pm 6.63 $ & $ 37.54 \pm 1.56 $ & $6.11$\\
    GloGNN &$88.31\pm1.13 $&$77.41\pm1.65 $&$89.62\pm0.35 $&$57.54\pm1.39 $&$69.78\pm2.42 $& $ 84.32  \pm 4.15$   &     $  87.06 \pm 3.53$       & $ 83.51 \pm 4.26 $ & $ 37.35 \pm 1.30 $ & $5.78$\\
    GloGNN++ &$ \underline{88.33\pm1.09} $&$77.22\pm 1.78$&$89.24\pm0.39 $&$57.88\pm1.76 $&$71.21\pm1.84 $&  $ 84.05  \pm 4.90$   &     $88.04 \pm 3.22$       & $ \underline{85.95 \pm 5.10}$ & $ \underline{37.70 \pm 1.40}$ & $4.00$\\
    ACM-GCN+ &$88.05\pm0.99 $&$\underline{77.67\pm 1.19}$ & $89.82\pm0.41$ & $\underline{66.98\pm1.71}$ & $\underline{74.47\pm1.84} $&  $ \mathbf{88.38  \pm 3.64}$   &     $\underline{88.43 \pm 2.39}$       & $ 85.68 \pm 4.84$ & $36.26 \pm 1.34$ & $3.78$ \\
    \hline
    \textbf{SADE-GCN} &$87.93\pm 0.91$ & $77.45\pm1.82$ & $\underline{90.07\pm0.46}$ & $\mathbf{68.20\pm1.57} $ & $\mathbf{75.57\pm1.57} $  & $86.49\pm5.12$ & $\mathbf{88.63\pm4.54}$& $\mathbf{86.21\pm5.59}$ & $\mathbf{37.91\pm0.97}$& $2.44$ \\
    \hline
    \end{tabular}
    }
\label{main-t1}
\end{table*}

\subsubsection{Baselines}
 The baseline models can be divided into four categories. The first category consists of MLP and LINK. MLP uses feature information exclusively, while LINK \cite{zheleva2009join} only incorporates topology information. The second category includes traditional GNNs, such as GCN \cite{kipf2016semi}, GAT \cite{velivckovicgraph}, GraphSAGE \cite{hamilton2017inductive}, MixHop \cite{abu2019mixhop}, and GCNII \cite{chen2020simple}. The third category encompasses GNNs designed for heterophily but does not explicitly encode topology information, such as H$_2$GCN \cite{zhu2020beyond}, FAGCN \cite{bo2021beyond}, GPR-GNN \cite{chien2020adaptive}, WRGAT \cite{li2022finding}, and ACM-GCN \cite{luanrevisiting}. The fourth category includes heterophilic graph neural networks that explicitly encode topology information, such as LINKX \cite{lim2021large}, GGCN \cite{yang2022graph}, GloGNN and GloGNN++ \cite{li2022finding}, and ACM-GCN+ \cite{luanrevisiting}. We obtain the baseline accuracies from \cite{li2022finding}, except for ACM-GCN+, which was not evaluated in \cite{li2022finding}. 

\subsubsection{Implementation Details}

For the nine small-scale datasets, we employ 10 public splits from \cite{peigeom}, reserving 48\% of nodes for training, 32\% for validation, and 20\% for testing in each split. The average test accuracy and standard deviation for each dataset are reported in our experiments. For the small graphs, we utilize SADE-GCN with two layers and 64 hidden dimensions. The cross-entropy loss is used for training SADE-GCN on the node classification task, and we optimize it using Adam \cite{kingma2014adam}. The complete hyperparameter settings can be found in the Appendix.

\subsection{Results}

\subsubsection{Node Classification}
We present the results of our method and other baselines on the node classification task in Table \ref{main-t1} and Table \ref{main-t2}. The results can be summarized as follows.

Traditional GNNs, such as GCN, GAT, and GraphSAGE, perform poorly on all heterophilic graphs and achieve better results on homophilic graphs. GCNII performs relatively well on heterophilic graphs and achieves state-of-the-art (SOTA) results on two homophilous graphs (Cora and PubMed) due to its initial residual and identity mapping.

The ACM family and the GloGNN family have achieved great success on both heterophilic and homophilous graphs, showing that negative attention weights can be helpful. The average ranking of heterophilic graph neural networks with explicitly encoded topology information is significantly better than those without. Although LINKX explicitly encodes topology and feature information, its pure MLP architecture lacks the ability to fully extract heterophily.

Our SADE-GCN achieves SOTA on graphs with varying levels of heterophily. We find that SADE-GCN performs better than MLP, LINK, and LINKX on every dataset, indicating that our method fully explores and combines feature and topology instead of compromising either.
Our model has a significant advantage over other methods for graphs dominated by topology, such as Squirrel, Chameleon. This is because our dual embedding architecture and innovative self-attention mechanism unleash the potential of topological information, helping to extract the heterophily of graphs. Our method outperforms FAGCN, which uses an additive attention mechanism, demonstrating the superiority of self-attention for graph data. However, all methods perform poorly on the Film dataset due to the poor quality of its feature and topology.

\subsubsection{Ablation Study}

\paragraph{the Importance of Feature and topology Information}

In this section, we analyze the impact of explicitly encoding the feature and topology information. We conducted ablation studies by removing either the feature embedding path $\mathbf{P}_{f}$ (W/o feature embedding) or the topology embedding path $\mathbf{P}_{t}$ (W/o topology embedding) in our SADE-GCN model. Our results indicate that SADE-GCN W/o feature performs better than SADE-GCN W/o topology on Cora, Squirrel and Chameleon, while the opposite result is observed for other datasets. This overall corresponds to the importance of feature and topology embedding shown by MLP and LINK.

\begin{table*}[!htbp]
\caption{Ablation study for SADE-GCN. }
    \centering
    \resizebox{\linewidth}{!}{
    \Huge
    \begin{tabular}{c c c c c c c c c c }
    \hline
    & \textbf{Cora} & \textbf{Citeseer}  & \textbf{Squirrel} & \textbf{Chameleon} & \textbf{Texas} & \textbf{Wisconsin}  & \textbf{Cornell} \\
    \hline
    \textbf{W/o feature embedding} &$ 83.76\pm1.22 $ & $ 68.78\pm2.95 $ &  $ 67.71\pm1.49 $ & $74.80\pm0.98 $  & $ 64.59\pm5.59 $ & $57.45\pm4.96$& $54.59\pm3.37$ &\\
    \textbf{W/o topology embedding} &$ 87.64\pm0.74 $ & $ 77.57\pm1.57 $ & $48.76\pm1.83 $ & $63.83\pm2.04 $  & $84.86\pm5.29$ & $86.66\pm3.01$& $85.67\pm4.84 $ & \\
    \hline
    \textbf{W/o feature scaling} & $ 86.27\pm1.53 $ & $ 76.77\pm1.53 $ & $68.07\pm1.55$ &$74.82\pm1.00 $  & $83.24\pm5.37$ & $84.90\pm4.02$& $84.86\pm6.64$ & \\
    \textbf{W/o topology scaling} &$ 87.38\pm0.66$ & $ 77.52\pm1.88$ & $ 68.00\pm1.81$ & $75.04\pm1.10$  & $ 81.62\pm7.43 $ & $85.68\pm5.95$& $85.13\pm5.57$&  \\
    \hline
    \textbf{SADE-GCN-sym} &$ 87.54\pm1.24 $ & $ 77.26\pm1.27 $ & $67.72\pm1.92 $ & $74.64\pm1.78 $  & $82.43\pm5.30$ & $85.88\pm3.48$& $83.51\pm6.66$ & \\
    \hline
    \textbf{SADE-GCN} &$87.93\pm 0.91$ & $77.45\pm1.82$ & $68.20\pm1.57$ & $75.57\pm1.57 $  & $86.49\pm5.12$ & $88.63\pm4.54$& $86.21\pm5.59$ &\\
    \hline
    \textbf{Graph Asymmetry} &$0.27$ & $0.22$ & $0.57$ & $0.40$  & $0.45$ & $0.39$& $0.39$ & \\
    \hline
    \end{tabular}
    }
\label{ab-all}
\end{table*}

\paragraph{the Importance of Scaling for Heterophily}
In this section, we explore the impact of feature and topology heterophily on the performance of SADE-GCN. The scaling function in equation \ref{e9} makes SADE-GCN have attention matrices with negative values to extract heterophily from feature and topology. We remove it from the embedding path $\mathbf{P}_{f}$, $\mathbf{P}_{t}$ for ablation. The result is in Table \ref{ab-all}, W/o feature ( topology) scaling corresponds to removing the scaling function from $\mathbf{P}_{f}$ ($\mathbf{P}_{t}$). we find that the absence of either will result in performance degradation. The more important the embedding is, the larger degradation when the scaling function is absent.

\paragraph{the Importance of Asymmetric Attention Matrix}

In this section, we analyze the asymmetric attention matrix of our model and propose a new variant of SADE, named SADE-GCN-sym, as a comparison. SADE-GCN-sym obtains a symmetric attention matrix through the inner product of the embeddings and their transposition. We introduce the concept of graph asymmetry, which quantifies the relative difference between the attention values of neighboring nodes. Specifically, the relative difference is defined as $ \frac{|\mathbf{\hat{R}}_{ij}-\mathbf{\hat{R}}_{ji}|}{|\mathbf{\hat{R}}_{ij}|+|\mathbf{\hat{R}}_{ji}|} $ for non-zero values. Graph asymmetry is computed as the average relative difference of attention matrices. We compare SADE-GCN-sym and graph asymmetry in Table \ref{ab-all}, and the results show that SADE-GCN outperforms SADE-GCN-sym on every dataset. This finding indicates that the asymmetric attention matrix is more effective. 

\subsubsection{the Robustness of Dual embedding Architecture} 

\begin{figure}[H]
    \centering
    \begin{minipage}{0.35\linewidth}
        \centering
        \includegraphics[width=0.95\linewidth]{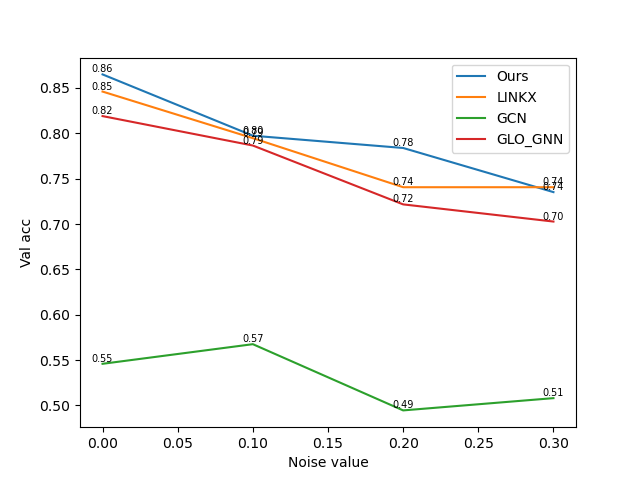}
         \subcaption{Feature embedding}
    \end{minipage}
    \begin{minipage}{0.35\linewidth}
        \centering
        \includegraphics[width=0.95\linewidth]{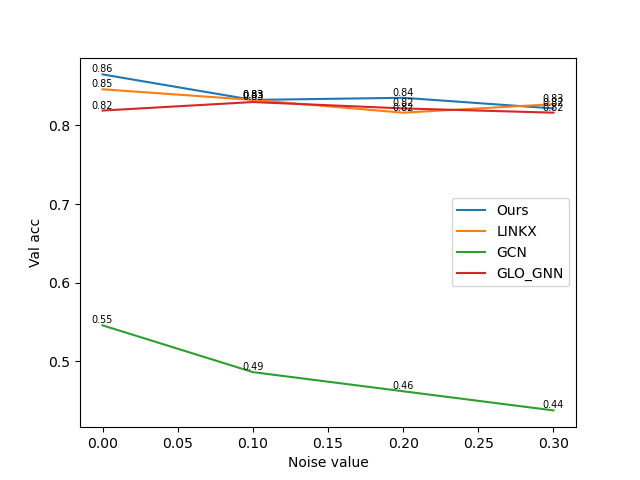}
        \subcaption{Topology embedding}
    \end{minipage}
    \caption{Embedding noise VS model accuracy (Texas) }
    \label{fig-noise}
\end{figure}

In this section, we conduct an experiment to verify the robustness of our dual embedding architecture. We add noise to the feature and topology respectively before training. For feature noise, we utilize the normal distribution and add it to the node feature using a coefficient range of $[0.0, 0.3]$. For topology noise, we randomly add or remove edges with a percentage range of $[0, 0.3]$ of the original number of edges. We compare our SADE-GCN with three representative models: LINKX, GloGNN, and GCN. LINKX embeds feature and topology separately using MLPs, concatenate the embeddings, and feed them to the final MLPs. GloGNN combines the feature and topology embeddings using a weighted sum in the first layer. We demonstrate the results of Texas in Figure \ref{fig-noise}. We found that our model achieves higher accuracy than the others for different levels of noise, showing that independently and explicitly embedding features and topology during message passing can make the model more robust. The feature noise is more influential in this case than the topology noise since Texas relies more on features. For graphs dominated by topology, the impact of topology noise is higher. The results of other graphs are presented in the Appendix.
\section{Related Work}
\label{sec:related}
\subsection{Graph Neural Network for Homophily and Heterophily}
 
Here, we introduce several representative GNNs. Traditional GNNs are effective on homophilic graphs, but not on heterophilic graphs. GCN performs graph convolution with a degree-normalized adjacency matrix and learnable transformation weight \cite{kipf2016semi}. GAT applies the attention mechanism to measure the importance of neighbor nodes during aggregation \cite{velivckovicgraph}. MixHop extends the aggregation to multi-hop neighbors for increased expressive ability \cite{abu2019mixhop}. Various GNNs have been designed to handle heterophilic graphs.
Geom-GCN aggregates topology and latent space neighbors to maintain topology information and long-range dependencies for heterophilic graphs \cite{peigeom}. H$_2$GNN is based on three design principles, including ego and neighbor separation, high-order neighbors, and multi-layer combination \cite{zhu2020beyond}. The weights contain positive and negative values, which help to learn both homophily and heterophily on the graph \cite{chien2020adaptive}. FAGCN first takes homophily and heterophily on the graphs as low and high frequency, respectively, and integrates different signals adaptively \cite{bo2021beyond}. GGCN adjusts degree coefficients and gives a signed message to jointly solve over-smoothing and heterophily problems \cite{yan2022two}. GloGNN and GloGNN++ use global feature similarity and topology similarity to conduct global message passing \cite{li2022finding}. ACM-GCN has three frequency channels (high, low, identity) to extract graph signals, and ACM-GCN+ combines the embeddings from frequency channels with topology embeddings \cite{luanrevisiting}.

\subsection{Graph Transformer}

Transformer has achieved great success on the graph task. There are several kinds of Graph Transformers(GT). 
GraphBERT\cite{zhang2020graph}, apply self-attention mechanism without aggregation. It samples subgraphs within their local contexts for training efficiency. It can be used as a pre-trained model for downstream tasks. Graphormer\cite{ying2021transformers} encodes the graph topology by novel, simple and effective methods. GPS\cite{rampavsek2022recipe} consists of positional and topology embedding, local message-passing, and global attention and keeps a balance between global and local information. 
GT has high parallelism and mitigates over-smoothing and over-squashing problems. In our method, we propose a novel self-attention mechanism inspired by GT to capture the heterophily and asymmetry of graphs.

\section{Conclusion}
In conclusion, we make three observations and corresponding analyses on heterophilic graphs. Firstly, we found that graph features and topology have varying degrees of importance and should be explicitly and independently embedded during message passing. Secondly, we discover that negative attention weights can be used to extract heterophily from both topology and feature embedding. Lastly, we identify the asymmetry present in heterophilic graphs. Based on these observations, we propose SADE-GCN, which explicitly encodes feature and topology through two independent paths and extracts heterophily and asymmetry through a novel self-attention mechanism. Our results show that SADE-GCN outperforms other models on heterophilic graphs and remains competitive on homophilic graphs. In future research, we plan to develop new sampling methods based on the heterophily and homophily of graphs to scale our model to larger graphs.

\clearpage

\bibliography{neurips_2023}
\bibliographystyle{plain}

\clearpage
\begin{appendices}

\section{Algorithm}
\subsection{Training Process of SADE-GCN}
We give an algorithm to summary the training process of SADE-GCN in Algorithm \ref{alg:SADE}. 
\begin{algorithm}[H]
    \caption{SADE-GCN Training}\label{alg:SADE}
    \begin{algorithmic}[1]
        \State \textbf{Input} { 
           $\mathbf{A}$, $\mathbf{X}$, $\mathbf{Y}_{\mathcal{T}}$, and a SADE-GCN model 
           $\mathbf{\emph{M}}$ with feature embedding path $\mathbf{P}_{f}$ and topology embedding path $\mathbf{P}_{t}$  
        }
        \For{$e$ = 1,2,...epochs}
            \For{$l$ = 1,2,...,$L$ }
                \If{$e$ \% $U$ == 0 }
                    \State  Update $\mathbf{\tilde{A}}_{f}^{(l)}$ for $\mathbf{P}_{f}$ and $\mathbf{\tilde{A}}_{t}^{(l)}$ for $\mathbf{P}_{t}$ in equation (4)(5)(6)(7). 
                \EndIf
                \State Do heterophilic graph convolution in $\mathbf{P}_{f}$ and $\mathbf{P}_{t}$ in equation (8)(9). 
            \EndFor
            \State  Adaptively combine feature and topology embeddings in equation (10).  
            \State  Calculate cross entropy loss for training. 
        \EndFor
    \end{algorithmic}
\end{algorithm}

Here is a time complexity analysis of SADE-GCN. 
The forward propagation of the network is from 2 to 11 lines of the above algorithm. We set the hidden dim as $H$. 
Updating $\mathbf{\tilde{A}}_{t}^{(l)}$ and $\mathbf{\tilde{A}}_{f}^{(l)}$ costs 
$\mathcal{O}(NH^{2}+N^{2}H)$. 
Do heterophilic graph convolution costs $\mathcal{O}(N^{2}H + NH^{2})$. 
Combining feature and topology costs $\mathcal{O}(NC^{2})$. 
Since there are $L$ layers, the total time complexity is $\mathcal{O}((NH^{2}+N^{2}H+NC^{2})L)$. 

\subsection{An Efficient Implementation for Graphs with Heterophily}

In most cases, the amount of graph nodes $N$ is much larger than the hidden dimension $H$ and the number of classes $C$. In those cases, the time and space complexity is dominated by the calculation of attention matrix.  The complexity of computing the attention matrix is quadratic in $N$. In order to make it possible for GPU training and speed up calculations, we give an efficient algorithm for computing the attention matrix. 

The implementation is as follows. 

\begin{algorithm}[H]
    \caption{Efficient Self-attention for Graphs with Heterophily}\label{alg:effi}
    \begin{algorithmic}[1]
        \State $Src(\mathbf{\tilde{A}})$ and $Dst(\mathbf{\tilde{A}}) $ represent the source node and destination node index of edges. 
        \State Arrange $\mathbf{Q}^{(l)}$ by $Src(\mathbf{\tilde{A}})$ to get $\mathbf{\hat{Q}}^{(l)}$ $\in \mathbb{R}^{E\times H^{(l)}}$, i.e. The $i$ row of $\mathbf{\hat{Q}}^{(l)}[i]$ corresponds to $\mathbf{Q}^{(l)}[Src(\mathbf{\tilde{A}})[i]]$ , for $i \in [1,E]$ . 
        \State $\mathbf{\hat{K}}^{(l)}$ $\in \mathbb{R}^{E\times H^{(l)}}$ can be obtained in In the same way above. 
        \State Do Hadamard product between $\mathbf{\hat{Q}}^{(l)}$ and $\mathbf{\hat{K}}^{(l)}$, and sum by row to get $\mathbf{R}^{(l)}$ stored in a sparse format.
        \State Use the scaling function as equation (6) to get the attention matrix $\mathbf{\hat{R}}^{(l)}$ with negative value. 
        \State Get the reweighted adjacency matrix $\mathbf{\tilde{A}}^{(l)}$ as equation (7). 
        
    \end{algorithmic}
\end{algorithm}
Algorithm \ref{alg:effi} shows that the complexity for updating $\mathbf{\tilde{A}}_{t}^{(l)}$ and $\mathbf{\tilde{A}}_{f}^{(l)}$ is $\mathcal{O}(NH^{2}+EH)$. The complexity for heterophilic graph convolution is $\mathcal{O}(EH + NH^{2})$. The total time complexity for SADE-GCN is $\mathcal{O}((NH^{2}+EH+NC^{2})L)$. Due to the fact that $E$ is much smaller than $N^{2}$ in most cases, the implementation can significantly reduce training time and memory usage.

\section{Dataset}
\textbf{Cora}, \textbf{Citeseer} and \textbf{Pubmed} are three citation graphs that commonly used as network benchmark datasets. In these graphs, nodes represent papers, and edges indicate the citation relationships between the paper. They use bag-of-words representations as the feature vectors of nodes. Each Node is given a label to represent the research field. Note that these datasets are homophilic graphs.

\textbf{Chameleon} and \textbf{Squirrel} are two Wikipedia pages on a specific topic. The nodes represent the web pages, edges are mutual links between pages. Node feature represents some informative nouns on the page. Nodes are divided into five categories based on monthly visits.  

\textbf{Cornell}, \textbf{Texas}, and \textbf{Wisconsin} datasets are collected by Carnegie Mellon University and comprise web pages as nodes and hyperlinks as edges to connect them. The node feature of these datasets is represented by the web page content in terms of their words. They are labeled into five categories, namely student, project, course, staff, and faculty.

\textbf{Film} is the subgraph selected from the film-director-actor-writer network. Nodes represent actor, and the edges denotes the co-occurence relationship on the same Wikipedia page. Node features are produced by the actors keywords in the Wikipedia pages. The task is to classify the actors into five categories.

\section{Hyper-parameters}

In our method, we utilize different types of attention function to combine the value and residual embedding in SAGC (combine-vr) and weight feature and topology embedding in the output layer (combine-ft). Both combine-vr and combine-ft offer three types of attention functions. The first type is the function used in equation (9) (type 1), while the second type is the adaptive attention method used in \cite{luanrevisiting} to combine different channel outputs (type 2). Finally, the third type lacks any attention function (type 3). For most graphs, $\alpha_{V}$ and $\alpha_{I}$ in equation (8) and $\alpha_{f}$ and $\alpha_{t}$ in equation (10) are set to 1. However, for Pubmed and Film datasets, setting $\alpha_{t}$ as zero during training results in better performance on the validation set. This indicates that on these datasets, topology can interfere with the expression of feature embedding and is not useful. For Film and twitch-gamer datasets, $\alpha_{V}$ is set to zero, indicating that the message-passing mechanism fails to perform on these graphs. Additionally, we can introduce non-linear layers and biases to $\mathbf{H}^{(l-1)}{\mathbf{W}_{I}}^{(l)}$ for better performance.

Next we introduce hyper-parameters we choose for training. For training process, the maximum epochs is in $ \left\{60, 220, 240, 300, 390, 500, 1800, 2100\right\}$. The searching range of learning rate is $ \left\{1e-3, 2e-3, 5e-3, 1.5e-2,2e-2,1e-2,5e-2, 6e-2\right\}$, the weight decay ranges in $ \left\{0.0, 1e-5, 2e-5, 5e-5, 1e-4, 2e-4, 5e-4, 1e-3, 2e-3, 5e-3, 1e-1 \right\}$. Dropout is in $ \left\{0.0, 0.1, 0.2, 0.3, 0.6,0.7,0.8,0.9\right\}$. $U$ in $\left\{1, 5, 10,20,50, 100, 240, 250\right\}$ is the epoch interval to update the attention matrix. For each training, we choose the model parameters with highest validation accuracy/ lowest validation loss. 

The hyper-parameters of SADE-GCN on each graph are shown as Table \ref{hp-t1}. 

\begin{table*}[!htbp]
\caption{The hyper-parameters of the nine small graphs.}
    \centering
    \resizebox{\linewidth}{!}{
    \Huge
    \begin{tabular}{c c c c c c c c  cc}
    \hline
    & \textbf{Cora} & \textbf{Citeseer} & \textbf{Pubmed} & \textbf{Squirrel} & \textbf{Chameleon} & \textbf{Texas} & \textbf{Wisconsin}  & \textbf{Cornell} & \textbf{Film} \\
    \hline
    combine-vr    & $2$ & $2$ & $2$ & $2$ & $1$ & $2$ & $2$ & $2$ & $2$ \\
    combine-ft    & $2$ & $1$ & $3$ & $3$ & $1$ & $2$ & $3$ & $2$ & $1$ \\
    $\alpha_{V}$  & $1$ & $1$ & $1$ & $1$ & $1$ & $1$ & $1$ & $1$ & $0$ \\ 
    $\alpha_{I}$  & $1$ & $1$ & $1$ & $1$ & $1$ & $1$ & $1$ & $1$ & $1$ \\ 
    $\alpha_{f}$  & $1$ & $1$ & $1$ & $1$ & $1$ & $1$ & $1$ & $1$ & $1$ \\ 
    $\alpha_{t}$  & $1$ & $1$ & $0$ & $1$ & $1$ & $1$ & $1$ & $1$ & $0$ \\
    non-linear    & $0$ & $0$ & $0$ & $0$ & $0$ & $0$ & $0$ & $0$ & $1$ \\  
    bias          & $0$ & $1$ & $0$ & $1$ & $0$ & $1$ & $1$ & $0$ & $1$ \\
    epochs        & $220$ & $60$ & $240$ & $1800$ & $300$ & $500$ & $2100$  & $390$ & $500$ \\
    dropout       & $0.6$ & $0.7$ & $0.3$ & $0.8$ & $0.8$ & $0.1$ & $0.1$ & $0.1$ & $0.0$ \\
    learning rate & $2e-3$ & $1e-2$ & $6e-2$ & $2e-3$  & $5e-2$ & $1.5e-2$  & $1.5e-2$ & $1e-2$ & $1e-2$ \\  
    weight decay  & $2e-4$ & $5e-5$ & $1e-5$ & $1e-4$ & $1e-4$ & $5e-4$ & $2e-4$  & $5e-4$ & $2e-3$  \\   
    $U$           & $5$ & $20$ & $240$ & $50$ & $10$ & $10$  & $5$ & $10$ & $500$ \\ 
    \hline
    \end{tabular}
    }
\label{hp-t1}
\end{table*}

\section{Experimental Results}

\subsection{Heterophily vs K-Hop Embedding Distance}
In this study, we present an experiment to demonstrate how our SADE-GCN model can extract graph heterophily from both feature and topology, highlighting the necessity of using a Self-attention mechanism to merge the embeddings.

By defining strict K-hop neighbors as nodes whose graph distances are equal to K, we can calculate the K-hop homophily, which is the ratio of K-hop neighbors sharing the same label. Additionally, we can measure the K-hop feature and topology embedding distance by computing the cosine distance\footnote{https://docs.scipy.org/doc/scipy/reference/generated/scipy.spatial.distance.cosine.html} between the feature and topology embeddings of a node and its K-hop neighbors, where the cosine distance serves as an effective similarity measure ranging from 0 to 2.

To showcase the effectiveness of our approach, we present results for each graph that indicate the K-hop homophily, K-hop feature embedding distance, and K-hop topology embedding distance. We represent feature and topology embeddings of nodes using $\mathbf{H}_{f}^{(L)}$ and $\mathbf{H}_{t}^{(L)}$, respectively. Note that for some small datasets like Wisconsin or Cornell, the high-order neighbor distance falls to 0. This is because no neighbor in this order for each of the nodes in the graph.

From the figures, we can make the following observations: 
Generally, all the figures follow the rule that when the homophily is high, the feature/structure distance is small, and vice versa. For homophily graphs like Cora and Citeseer, The farther away the neighbors are, the greater the embedding distance. For other heterophily graphs, the distance changes with homophily. This is reasonable because high homophily indicates neighbor nodes tend to be in the same class, so their feature/topology similarity should be higher, and the distance is smaller. These results also show that our model can mine into intrinsic heterophily of datasets.

\begin{figure}[H]
    \centering
    \vspace{-0.4cm}
    \begin{minipage}{0.3\linewidth}
        \centering
        \includegraphics[width=0.95\linewidth]{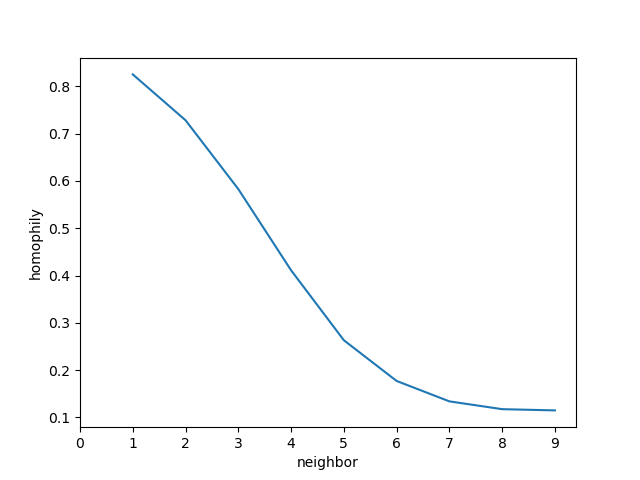}
         \subcaption{Homophily}
    \end{minipage}
    \begin{minipage}{0.3\linewidth}
        \centering
        \includegraphics[width=0.95\linewidth]{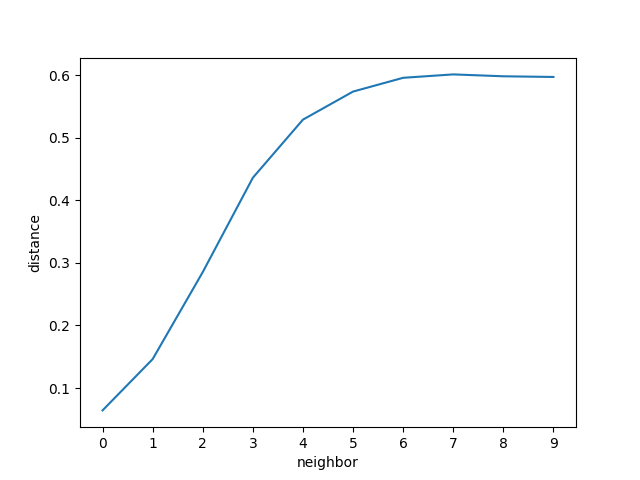}
         \subcaption{Feature distance}
    \end{minipage}
    \begin{minipage}{0.3\linewidth}
        \centering
        \includegraphics[width=0.95\linewidth]{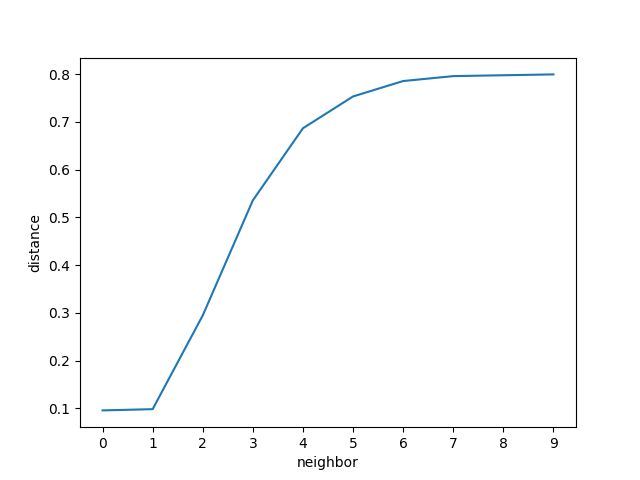}
        \subcaption{Topology distance}
    \end{minipage}
    \caption{Homophily and Feature/Topology embedding distance (Cora) }
    \label{fig-dist-cora}
\end{figure}
\begin{figure}[H]
    \centering
    \vspace{-0.4cm}
    \begin{minipage}{0.3\linewidth}
        \centering
        \includegraphics[width=0.95\linewidth]{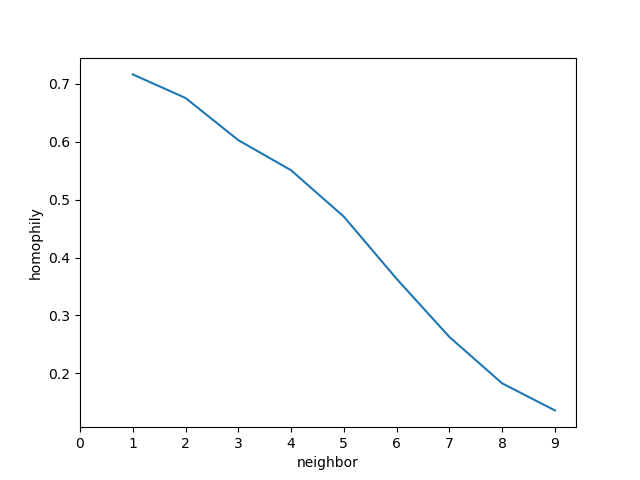}
         \subcaption{Homophily}
    \end{minipage}
    \begin{minipage}{0.3\linewidth}
        \centering
        \includegraphics[width=0.95\linewidth]{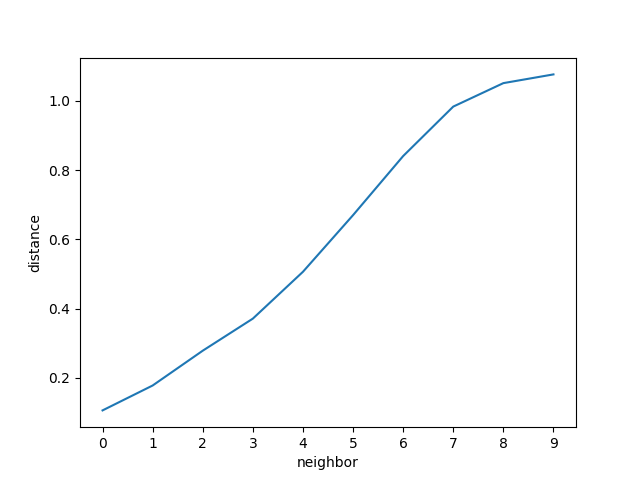}
         \subcaption{Feature distance}
    \end{minipage}
    \begin{minipage}{0.3\linewidth}
        \centering
        \includegraphics[width=0.95\linewidth]{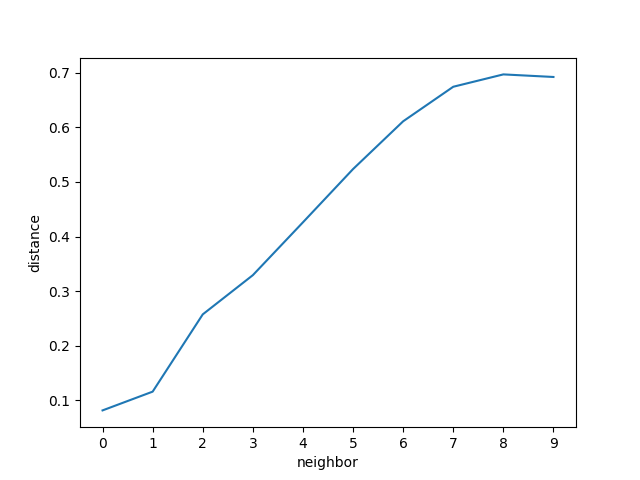}
        \subcaption{Topology distance}
    \end{minipage}
    \caption{Homophily and Feature/Topology embedding distance (Citeseer) }
    \label{fig-dist-citeseer}
\end{figure}
\begin{figure}[H]
    \centering
    \vspace{-0.4cm}
    \begin{minipage}{0.3\linewidth}
        \centering
        \includegraphics[width=0.95\linewidth]{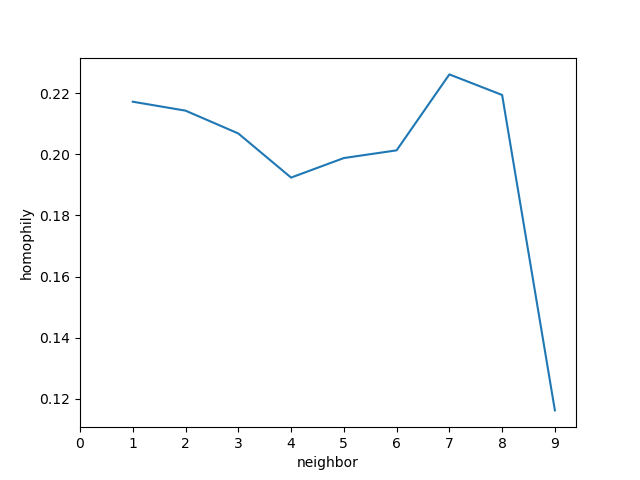}
         \subcaption{Homophily}
    \end{minipage}
    \begin{minipage}{0.3\linewidth}
        \centering
        \includegraphics[width=0.95\linewidth]{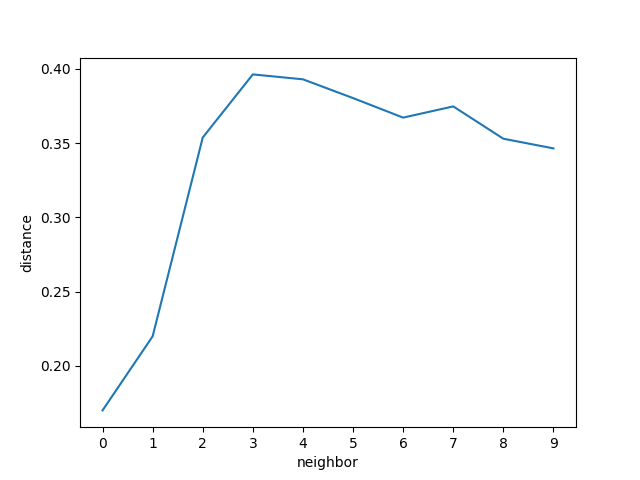}
         \subcaption{Feature distance}
    \end{minipage}
    \begin{minipage}{0.3\linewidth}
        \centering
        \includegraphics[width=0.95\linewidth]{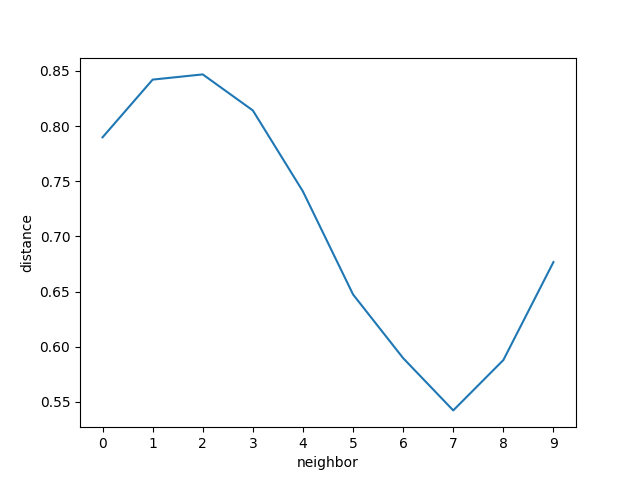}
        \subcaption{Topology distance}
    \end{minipage}
    \caption{Homophily and Feature/Topology embedding distance (Squirrel) }
    \label{fig-dist-squirrel}
\end{figure}
\begin{figure}[H]
    \centering
    \vspace{-0.4cm}
    \begin{minipage}{0.3\linewidth}
        \centering
        \includegraphics[width=0.95\linewidth]{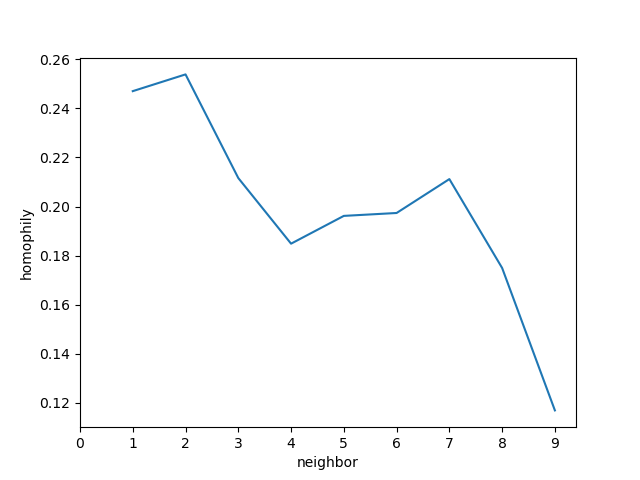}
         \subcaption{Homophily}
    \end{minipage}
    \begin{minipage}{0.3\linewidth}
        \centering
        \includegraphics[width=0.95\linewidth]{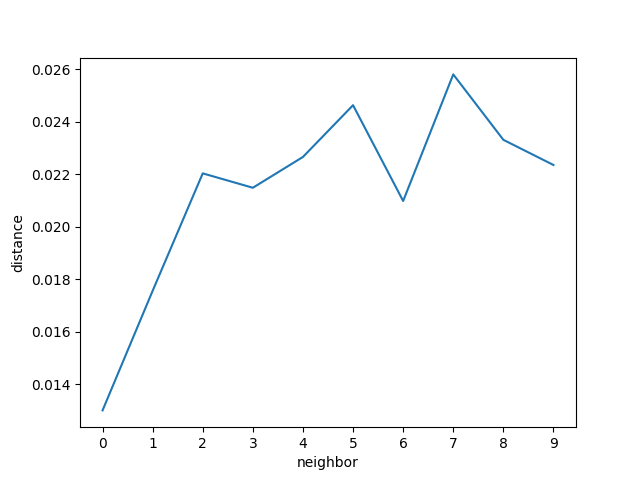}
         \subcaption{Feature distance}
    \end{minipage}
    \begin{minipage}{0.3\linewidth}
        \centering
        \includegraphics[width=0.95\linewidth]{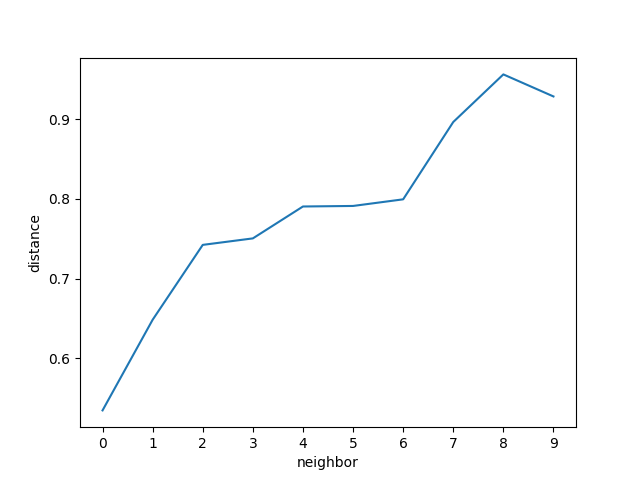}
        \subcaption{Topology distance}
    \end{minipage}
    \caption{Homophily and Feature/Topology embedding distance (Chameleon) }
    \label{fig-dist-chameleon}
\end{figure}
\begin{figure}[H]
    \centering
    \vspace{-0.4cm}
    \begin{minipage}{0.3\linewidth}
        \centering
        \includegraphics[width=0.95\linewidth]{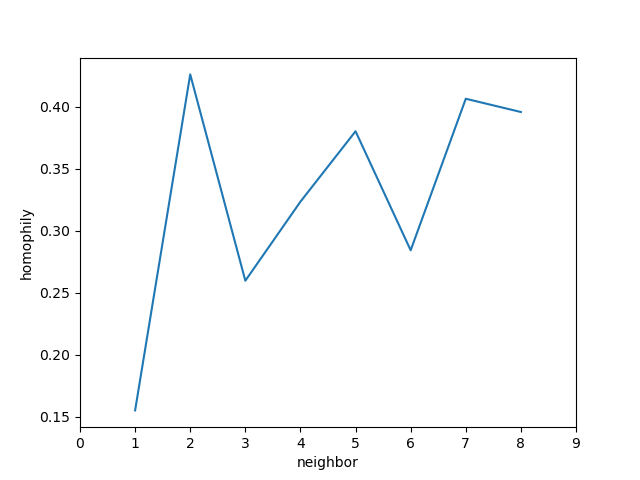}
         \subcaption{Homophily}
    \end{minipage}
    \begin{minipage}{0.3\linewidth}
        \centering
        \includegraphics[width=0.95\linewidth]{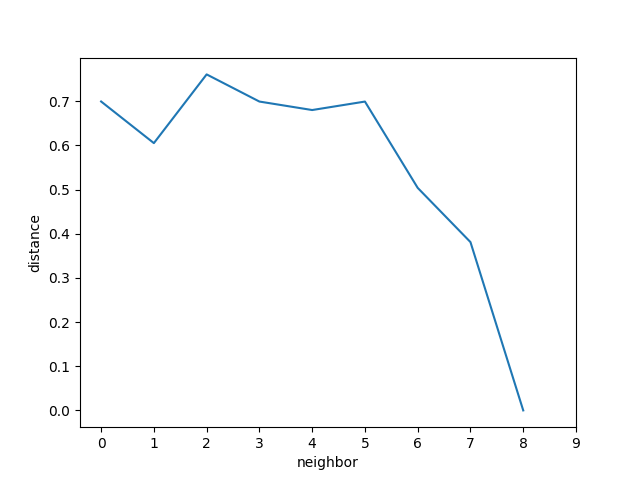}
         \subcaption{Feature distance}
    \end{minipage}
    \begin{minipage}{0.3\linewidth}
        \centering
        \includegraphics[width=0.95\linewidth]{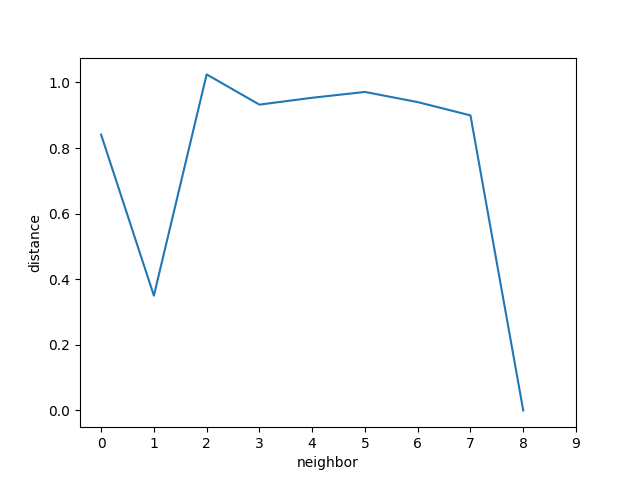}
        \subcaption{Topology distance}
    \end{minipage}
    \caption{Homophily and Feature/Topology embedding distance (Wisconsin) }
    \label{fig-dist-wisconsin}
\end{figure}
\begin{figure}[H]
    \centering
    \vspace{-0.4cm}
    \begin{minipage}{0.3\linewidth}
        \centering
        \includegraphics[width=0.95\linewidth]{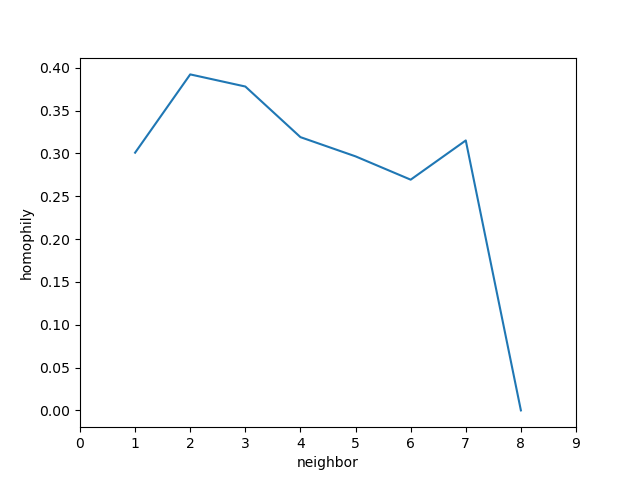}
         \subcaption{Homophily}
    \end{minipage}
    \begin{minipage}{0.3\linewidth}
        \centering
        \includegraphics[width=0.95\linewidth]{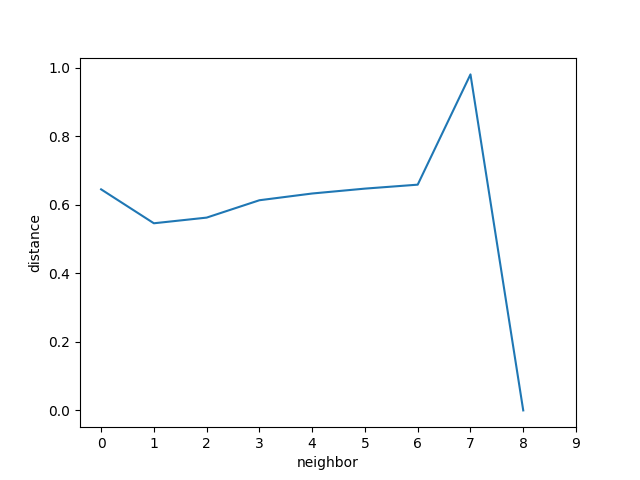}
         \subcaption{Feature distanceg}
    \end{minipage}
    \begin{minipage}{0.3\linewidth}
        \centering
        \includegraphics[width=0.95\linewidth]{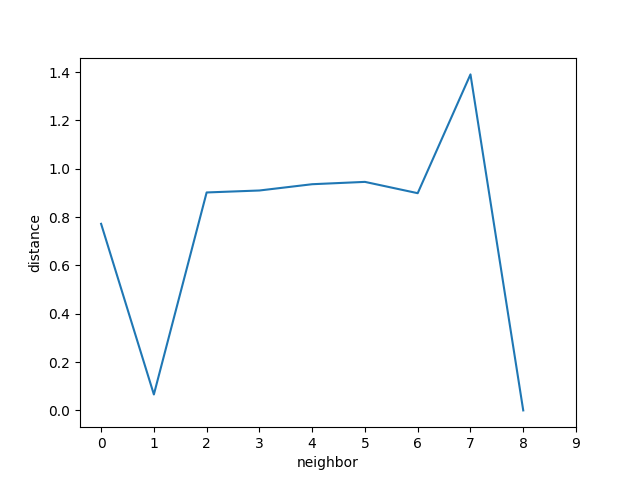}
        \subcaption{Topology distance}
    \end{minipage}
    \caption{Homophily and Feature/Topology embedding distance (Cornell) }
    \label{fig-dist-cornell}
\end{figure}
\begin{figure}[H]
    \centering
    \vspace{-0.4cm}
    \begin{minipage}{0.3\linewidth}
        \centering
        \includegraphics[width=0.95\linewidth]{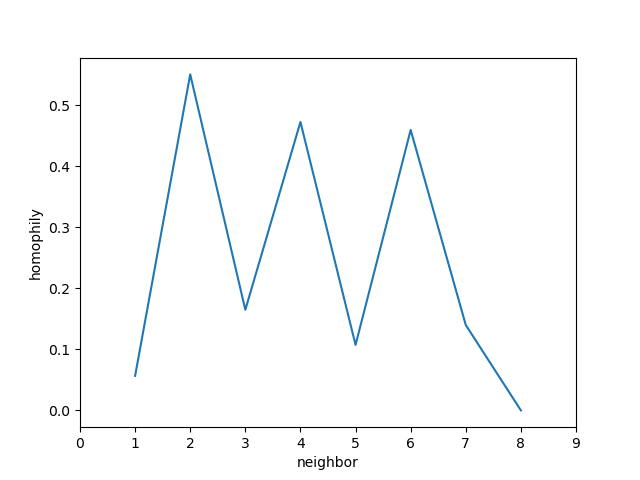}
         \subcaption{Homophily}
    \end{minipage}
    \begin{minipage}{0.3\linewidth}
        \centering
        \includegraphics[width=0.95\linewidth]{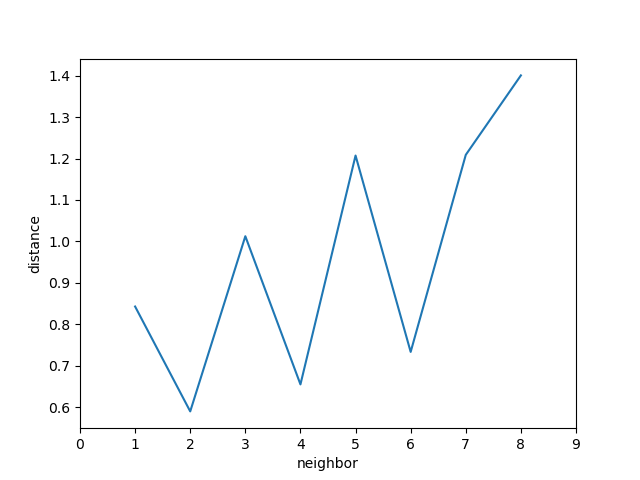}
         \subcaption{Feature distance}
    \end{minipage}
    \begin{minipage}{0.3\linewidth}
        \centering
        \includegraphics[width=0.95\linewidth]{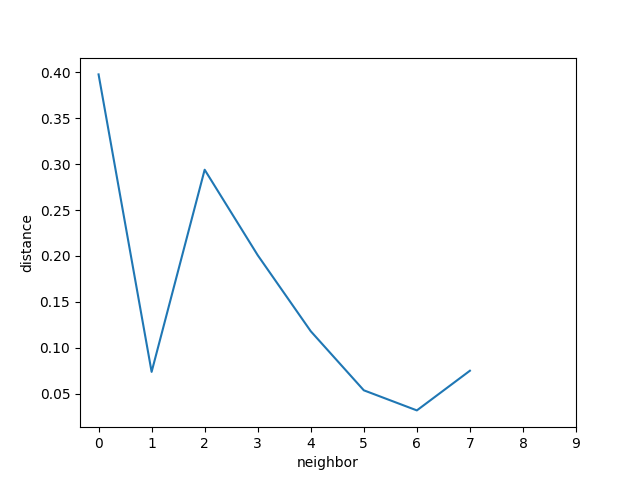}
        \subcaption{Topology distance}
    \end{minipage}
    \caption{Homophily and Feature/Topology embedding distance (Texas) }
    \label{fig-dist-texas}
\end{figure}
\subsection{the Robustness of Dual Embedding Architecture}
\begin{figure}[H]
    \centering
    \vspace{-0.4cm}
    \begin{minipage}{0.35\linewidth}
        \centering
        \includegraphics[width=0.95\linewidth]{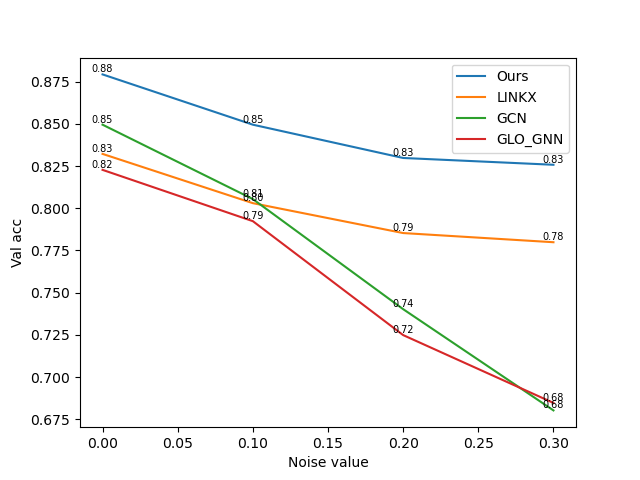}
         \subcaption{Feature embedding}
    \end{minipage}
    \begin{minipage}{0.35\linewidth}
        \centering
        \includegraphics[width=0.95\linewidth]{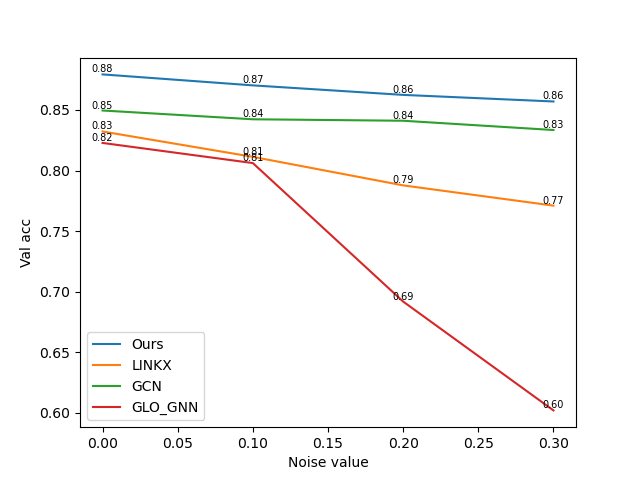}
        \subcaption{Topology embedding}
    \end{minipage}
    \caption{Embedding noise VS model accuracy (Cora) }
    \label{fig-noise-cora}
\end{figure}

\begin{figure}[H]
    \centering
    \vspace{-0.4cm}
    \begin{minipage}{0.35\linewidth}
        \centering
        \includegraphics[width=0.95\linewidth]{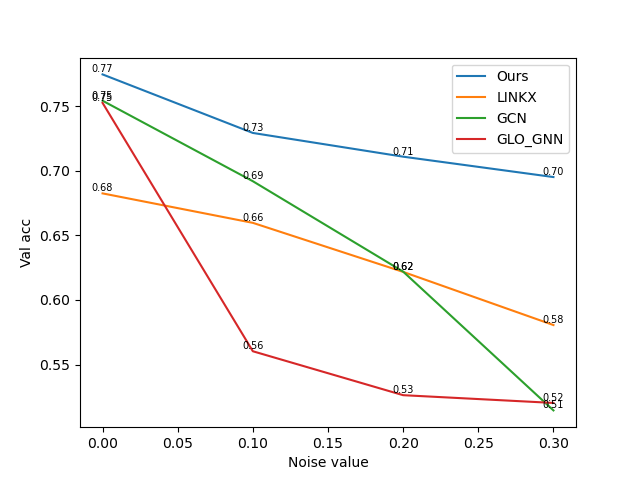}
         \subcaption{Feature embedding}
    \end{minipage}
    \begin{minipage}{0.35\linewidth}
        \centering
        \includegraphics[width=0.95\linewidth]{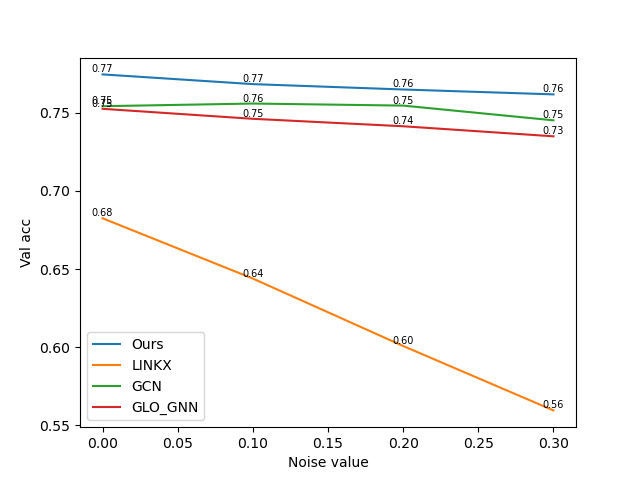}
        \subcaption{Topology embedding}
    \end{minipage}
    \caption{Embedding noise VS model accuracy (Citeseer) }
    \label{fig-noise-citeseer}
\end{figure}

\begin{figure}[H]
    \centering
    \vspace{-0.4cm}
    \begin{minipage}{0.35\linewidth}
        \centering
        \includegraphics[width=0.95\linewidth]{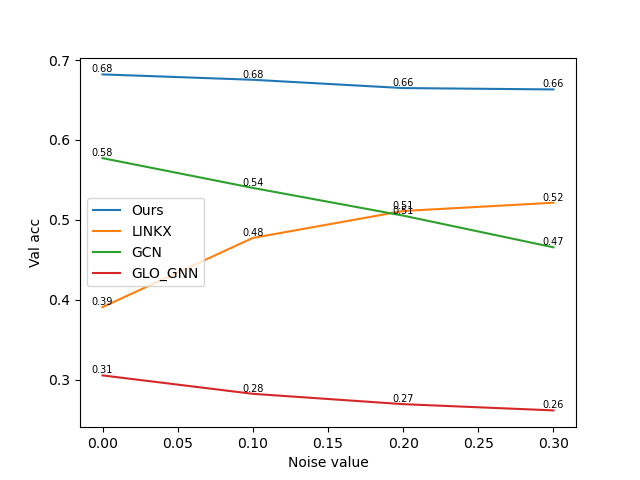}
         \subcaption{Feature embedding}
    \end{minipage}
    \begin{minipage}{0.35\linewidth}
        \centering
        \includegraphics[width=0.95\linewidth]{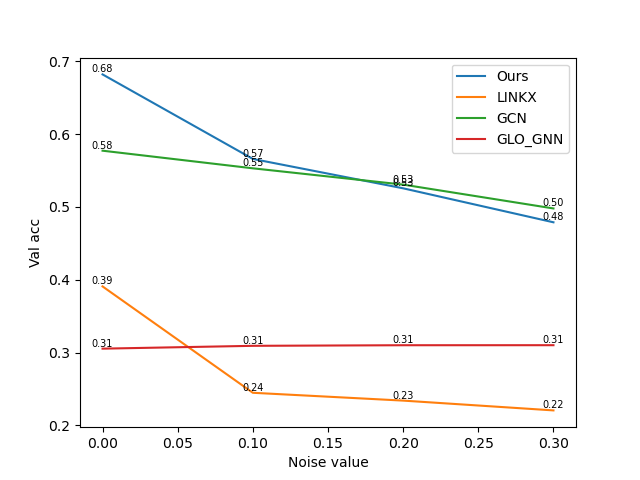}
        \subcaption{Topology embedding}
    \end{minipage}
    \caption{Embedding noise VS model accuracy (Squirrel) }
    \label{fig-noise-squirrel}
\end{figure}

\begin{figure}[H]
    \centering
    \vspace{-0.4cm}
    \begin{minipage}{0.35\linewidth}
        \centering
        \includegraphics[width=0.95\linewidth]{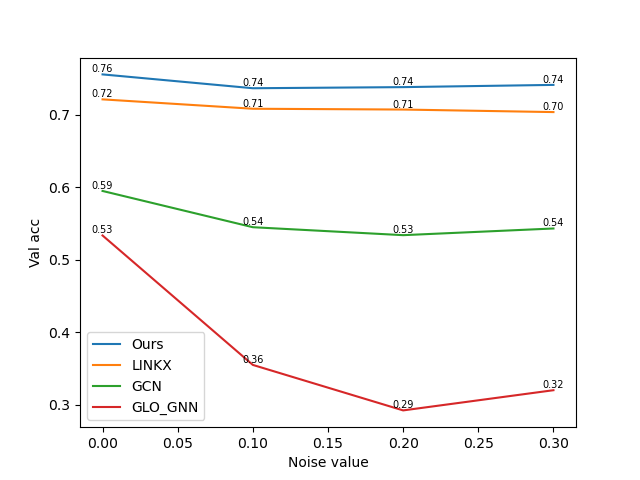}
         \subcaption{Feature embedding}
    \end{minipage}
    \begin{minipage}{0.35\linewidth}
        \centering
        \includegraphics[width=0.95\linewidth]{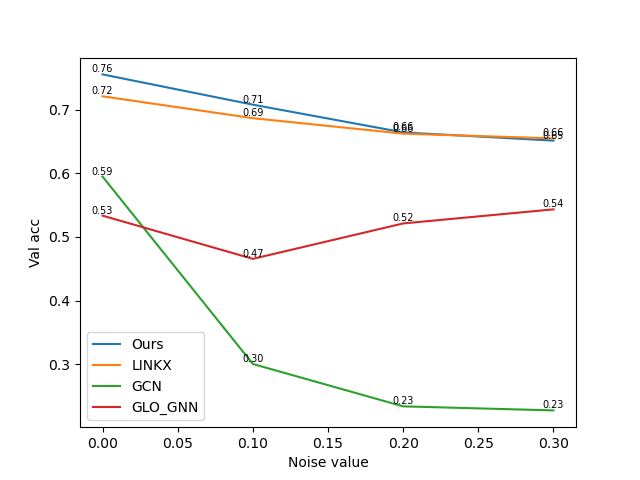}
        \subcaption{Topology embedding}
    \end{minipage}
    \caption{Embedding noise VS model accuracy (Chameleon) }
    \label{fig-noise-chameleon}
\end{figure}

\begin{figure}[H]
    \centering
    \vspace{-0.4cm}
    \begin{minipage}{0.35\linewidth}
        \centering
        \includegraphics[width=0.95\linewidth]{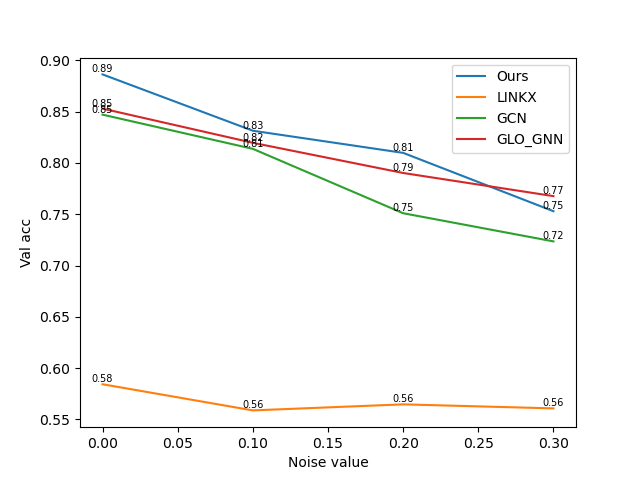}
         \subcaption{Feature embedding}
    \end{minipage}
    \begin{minipage}{0.35\linewidth}
        \centering
        \includegraphics[width=0.95\linewidth]{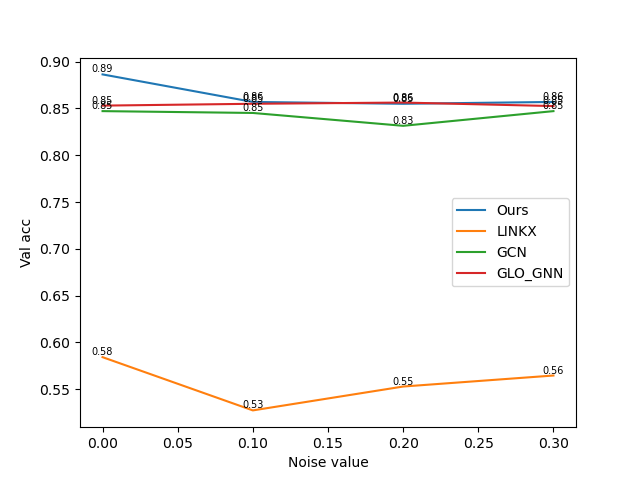}
        \subcaption{Topology embedding}
    \end{minipage}
    \caption{Embedding noise VS model accuracy (Wisconsin) }
    \label{fig-noise-wisconsin}
\end{figure}

In Section 4.2.3, we demonstrate an experiment to verify the robustness on Texas. Here is the results of other graphs for the experiment.  The Figures \ref{fig-noise-cora}\ref{fig-noise-citeseer}\ref{fig-noise-squirrel}\ref{fig-noise-chameleon}\ref{fig-noise-wisconsin} thoroughly verify the arguments we put forward, i.e. independently and explicitly embedding feature and topology during message passing can make the model more robust. 
\begin{figure}[H]
    \centering
    \vspace{-0.4cm}
    \begin{minipage}{0.35\linewidth}
        \centering
        \includegraphics[width=0.95\linewidth]{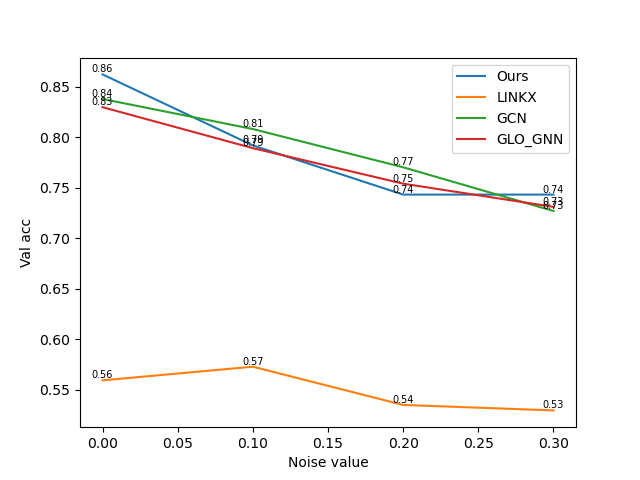}
         \subcaption{Feature embedding}
    \end{minipage}
    \begin{minipage}{0.35\linewidth}
        \centering
        \includegraphics[width=0.95\linewidth]{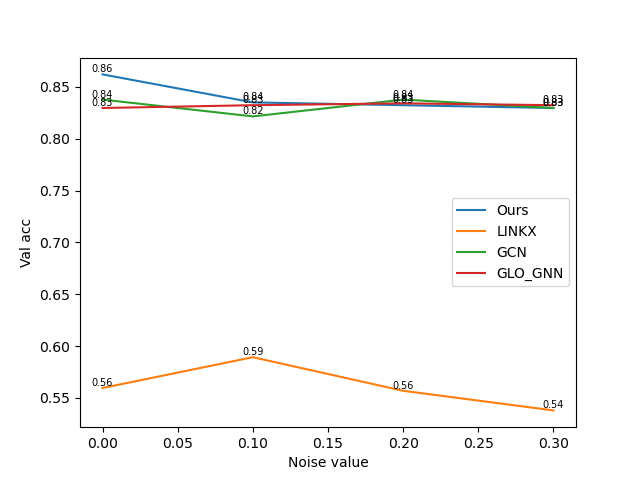}
        \subcaption{Topology embedding}
    \end{minipage}
    \caption{Embedding noise VS model accuracy (Cornell) }
    \label{fig-noise-cornell}
\end{figure}

\section{Code}

 Our code will be available soon. For data preprocessing, we refer to GloGNN\cite{li2022finding} and ACM-GNN\cite{luanrevisiting}. Their code can be found in \footnote{https://github.com/RecklessRonan/GloGNN} and \footnote{https://github.com/SitaoLuan/ACM-GNN}. 

\end{appendices}

\end{document}